\documentclass[10pt,journal,compsoc]{IEEEtran}

\usepackage[nocompress]{cite}
\usepackage{amsmath}
\usepackage{booktabs}
\usepackage{amssymb}
\usepackage{nicefrac}
\usepackage{algorithm} 
\usepackage{algorithmic} 
\usepackage{enumitem}
\usepackage{graphicx}
\usepackage[table,xcdraw]{xcolor}
\usepackage[caption=false,font=footnotesize,labelfont=sf,textfont=sf]{subfig}
\usepackage{dblfloatfix}
\usepackage{flushend}
\usepackage{url}

\definecolor{sig-green}{HTML}{79de79}
\definecolor{sig-blue}{HTML}{a8e4ef}
\definecolor{sig-yellow}{HTML}{fcfc99}
\definecolor{sig-red}{HTML}{fb6962}

\begin{document}

\title{Generalized Reciprocal Perspective}

\author{Kevin~Dick,~\IEEEmembership{Graduate~Student~Member,~IEEE,}
 Daniel~G.~Kyrollos,~\IEEEmembership{Graduate~Student~Member,~IEEE}
 and~James~R.~Green,~\IEEEmembership{Senior~Member,~IEEE}
\IEEEcompsocitemizethanks{\IEEEcompsocthanksitem K. Dick, D. G. Kyrollos, and J. R. Green are  with the Department of Systems \& Computer Engineering, Carleton University, Ottawa, Ontario, Canada.\protect\\
E-mail: \{kevin.dick,daniel.kyrollos,james.green\}@carleton.ca

\IEEEcompsocthanksitem The authors are also members of the Institute of Data Science, Carleton University, Ottawa, Canada.}
\thanks{}}

\markboth{}{Dick \MakeLowercase{\textit{et al.}}: Generalized Reciprocal Perspective}

\IEEEtitleabstractindextext{%
\begin{abstract}

Across many application domains, real-world problems can be represented as a network, with nodes representing domain-specific elements and the edges (or lack thereof) capturing the relationship between elements. Two notable example domains that leverage link prediction algorithms are the tasks of protein-protein interaction (PPI) prediction, in the bioinformatic domain, and the prediction of user-item five-point scale rating, in the eCommerce domain. Leveraging high-performance computing and optimized link prediction algorithms, it is increasingly possible to evaluate every possible combination of nodal pairs enabling the generation of a comprehensive prediction matrix (CPM) that places an individual link prediction score in the context of all possible links involving either node (providing data-driven \textit{context}). Historically, this contextual information has been ignored given exponentially growing problem sizes resulting in computational intractability; however, we demonstrate that expending high-performance compute resources to generate CPMs is a worthwhile investment given the improvement in predictive performance. In this work, we generalize for all pairwise link-prediction tasks our novel semi-supervised machine learning method, denoted Reciprocal Perspective (RP). We demonstrate that RP significantly improves link prediction accuracy by leveraging the wealth of information in a CPM. Context-based features are extracted from the CPM for use in a stacked classifier that refines link prediction scores. Herein, we leverage three independent data sources (MovieLens, GoodBooks, Amazon Product Categories) containing 5-point ratings data across 24 different item categories and differing user-bases and 13 recommendation system (RS) algorithms (including an equi-weighted ensemble of all 12 component RS algorithms). We demonstrate that the application of RP in a cascade almost always results in significantly ($p < 0.05$) improved predictions. These results on RS-type problems, combined with previously published performance on both homogeneous and heterogeneous link prediction, in both classification- and regression-task formulations, suggest that RP is applicable to a broad range of link prediction problems and we recommend its use in any pairwise link prediction pipeline for which the CPM may be generated. The RP framework is available for use from the following GitHub repository: \url{https://github.com/GreenCUBIC/RP}.


\end{abstract}

}

\maketitle
\IEEEdisplaynontitleabstractindextext
\IEEEpeerreviewmaketitle

\IEEEraisesectionheading{\section{Introduction}\label{sec:introduction}}

\IEEEPARstart{I}{nnumerable} systems across diverse fields of science and engineering can be abstracted into networks of interconnected elements. These networks conventionally comprise a set of nodes representing individual and distinct elements, and a set of edges\footnote{In this work we use the terms \textit{edge} and \textit{link} interchangeably as well as the terms \textit{network} and \textit{graph}.} which occur between pairs of nodes. Networks are flexible and abstract models that can represent diverse concepts. Popularized examples of element-element relationships in networks include social networks (nodes: individuals; edges: social tie), telecommunication systems (nodes: terminals; edges: transmission links), and biological systems (nodes: proteins; edges: molecular interactions). Indeed, advances within the domain of Network Science reverberate through many disparate domains having wide bearing on the creation and understanding of those complex systems. 

\subsection{Pairwise Link Prediction in Network Science}

Network Science as an academic field is methodologically diverse and ever-evolving \cite{lewis2011network}. Notably, the field draws from graph theory in mathematics \cite{van2010graph}, social structure from sociology \cite{cordeiro2018evolving}, inferential modelling from statistics \cite{handcock2003statistical}, and statistical mechanics from physics \cite{albert2002statistical}. This work and the methods described herein are contributions to link prediction and data mining from computer science \cite{lu2011link}.

Link prediction is a fundamental problem in modern information science. Certain traditional approaches leveraged methods such as Markov models and/or statistical modeling to infer new edges \cite{wang2005group, getoor2007markov}; however, these fail to capture structural characteristics of networks such as communities, hubs, and hierarchical organization \cite{getoor2005link}. It is the nascence of the Netflix competition \textit{circa.} 2006-2009 that spurred research engagement and the popularization of similarity-based and matrix-factorization type link predictors, notably for the task of predicting user-movie ratings on the five-point star scale \cite{bell2007lessons}. Through the  following decade and with the emergence of deep learning and the growing availability of high-performance computing infrastructure, new classes of network representations have emerged including \textit{network embedding} that seeks to map higher-dimensional nodes to a lower dimensional latent space while conserving neighbourhood structure \cite{kazemi2018simple}. Two notable examples include DeepWalk \cite{perozzi2014deepwalk} and \verb|node2vec| \cite{grover2016node2vec}, each with demonstrated improvement of link prediction performance for diverse domain applications. Beyond graph representation learning, an emergent class of end-to-end graph neural networks (GNN) have been introduced \cite{zhang2018end} enabling improved graph feature learning \cite{zhang2018link}. (Deep) GNNs are reported to be highly competitive with state-of-the-art (SOTA) graph kernel methods and outperform deep learning methods for graph classification tasks \cite{zhang2018end}. 

\subsection{Pairwise Relationships for Scientific Discovery}

Accurate link prediction for applied real-world problems have important consequence to the understanding and interpretation of the complex systems that they represent. Algorithms capable of accurately predicting missing links enable data mining applications, accelerate network data collection, and improve network model validation. Recent trends have seen improved performance in link predictions through multi-sided recommendation, model calibration, model stacking, and/or large-scale ensembles.

In the work of Berlusconi \textit{et al.}, link prediction was leveraged to identify possible missing links in a criminal network by considering multi-sided similarity measures of pairs of nodes in the network and inferring them \textit{a contrario} with the assumption that putative social ties will be characterized by \textit{opposite} features \cite{berlusconi2016link}.

Model ``stacking" is an ensemble approach that learns a meta-model that learns how to leverage the predictions from individual component predictors \cite{wolpert1992stacked} that differs from conventional bagging and boosting. Unlike bagging, in stacking, the contributing component models are typically diverse (\textit{e.g.} a variety of algorithms or (deep) learning models) and fit on the same dataset (as opposed to a sampling of the training dataset). Unlike boosting, in stacking, a single model is used to learn how to best combine the predictions from the contributing models as opposed to a sequence of models that correct the predictions of prior models.

Ghasemian \textit{et al.} reported a systematic evaluation of 203 individual link predictor algorithms, representing three popular families of methods, applied to a large corpus of 550 structurally diverse networks from six scientific domains \cite{ghasemian2020stacking}. Excitingly, the stacked models achieved (near) optimal levels of accuracy over synthetically generated datasets for which the maximally achievable level of performance was known and the stacked meta-classifiers classifiers, when trained on real-world datasets, were consistently superior to component models \cite{ghasemian2020stacking}. These findings demonstrate the broad utility of stacked meta-classification methods on diverse problem sets.

Finally, model calibration is crucial in high-stakes scenarios such as drug-target interaction (DTI) prediction where end-users need trustworthy and interpretable decisions. In a binary classification formulation (\textit{e.g.} positive prediction indicates a putative interaction and a negative prediction indicates non-interaction) probability calibration is important when the confidence in a given prediction must make probabilistic sense. For example, if a given model predicts a fact is true with 80\% confidence, the model should be correct 80\% of the time. Adherence to this property is evaluated by means of calibration/reliability curves which plot the model's mean predicted value along the x-axis and the fraction of positives along the y-axis with the identity function representing perfect calibration. In the work of Tabacof and Costabello, the application of Platt scaling \cite{platt1999probabilistic} and isotonic regression \cite{zadrozny2002transforming} was used to calibrate knowledge graph embedding models \cite{tabacof2019probability} and Wang \textit{et al.} proposed methods broadly applicable to graph neural networks \cite{wang2021confident}. Calibration methods are a form of post-processing/model stacking for refining prediction scores, which is conceptually similar to RP.

\subsection{Reciprocal Perspective for Biomedical Discovery}

The fields of bioinformatics and computational biology have been central application domains for the study and exemplification of network-based methods; these approaches have been widely used to investigate biological systems at various scales, whether in macroscopic ecological dynamics down to microscopic and molecular interaction studies \cite{gosak2018network}. The Reciprocal Perspective (RP) methodology was first discovered and investigated within these domains by reframing several applications as pairwise link prediction problems.

Briefly, the RP framework is a cascaded classifier that refines raw pairwise link prediction scores by considering the context of all possible link scores involving either element of the pair. To provide a concrete example, consider the task of predicting all protein-protein interactions (PPI) among an organism's $n$ proteins. From all the \nicefrac{n(n+1)}{2} possible pairs of proteins, a subset are known to interact (positive) or not interact (negative) through experimental validation studies. These known links are useful for training and evaluating a PPI prediction algorithm, denoted $f_{\Theta}(x)$. RP begins by applying this \textit{initial predictor}, $f_{\Theta}(x)$, to infer prediction scores between all \nicefrac{n(n+1)}{2} possible pairs (including those that are known via a cross-validation schema). This results in a complete $K_n$ weighted-edged graph of predicted scores from $f_{\Theta}(x)$ with a corresponding complete adjacency matrix of predicted scores that we denote the Comprehensive Prediction Matrix (CPM). A given row $i$ sliced from the matrix is a $n \times 1$ vector of all scores between protein $i$ and every other protein in the proteome. Similarly, a given column $j$ sliced from the matrix is a $1 \times n$ vector of all the scores between protein $j$ and every other protein in the proteome (including protein $i$ in cell $i,j$). These vectors, when each sorted in rank-order by monotonically decreasing score, represent a pair of \textit{One-to-All} (O2A) score curves. The pair of O2A curves share only a single common value representing the query pair $i,j$ (identical in value, but usually differing in sorted rank). These O2A curves (which we also refer to as protein $i$'s \textit{perspective}, and protein $j$'s \textit{perspective)} typically exhibit a characteristic "S"- or "L"-shaped distribution with a  \textit{baseline} that we can attribute to non-interacting pairs. The singular common point between the two reciprocal perspectives enable a number of numeric features to be computed characterizing the location of that point in the broader context of all the inferred scores within the two distributions and with respect to their baselines. Intuitively, we wish to identify a pair for which their shared score would be relatively high-scoring with respect to each perspective's baseline. Thus, the RP framework extracts, for any pair of proteins $i,j$, a \textit{new} numerical vector of O2A-derived features that can be leveraged to subsequently train and evaluate a cascaded predictor that refines the original predictions. Additionally, the cascaded predictor can also function as a means of combining multiple experts (CME) by fusing the RP feature vectors obtained from the CPMs generated by numerous initial predictors to function in a \textit{stacked generalization} schema.

The original RP formulation was proposed for intra-species PPI prediction tasks \cite{dick2018reciprocal} and then later for improving inter- and cross-species SOTA predictors \cite{dick2020pipe4}. In each case, the link prediction problems were formulated as a binary classification task seeking to maximize F1 score, precision, and recall. Thereafter, RP demonstrated significant improvement for the task of micro-RNA (mRNA) target prediction in \cite{kyrollos2020rpmirdip}. This work formulated the link prediction task as a classification-type problem seeking to maximize the Area under the Reciever Operating Characteristic curve (AUC ROC) and the Area under the Precision-Recall curve (AUC PR). Most recently, the RP framework was leveraged for drug-target interaction prediction \cite{musdti}; RP was here generalized to regression-type problems to minimize the Root Mean Squared Error (RMSE) and maximize the Concordance Index (CI). Finally, the original PPI problem formulation was leverage for the study of SARS-CoV-2 \cite{dick2021multi} and adapted to use the cascaded RP model for the CME in a $n=2$ stacking of two SOTA PPI classifiers. In each problem domain, the biological elements being operated upon were arranged into a network-like representation. This suggests that the RP framework may be effectively generalized to operate on abstracted element-element link prediction in the broader domain of Network Science.

\subsection{Generalizing Reciprocal Perspective for (m)any Pairwise Applications}

\begin{figure*}[bt]
  \centering
  \includegraphics[width=0.9\textwidth]{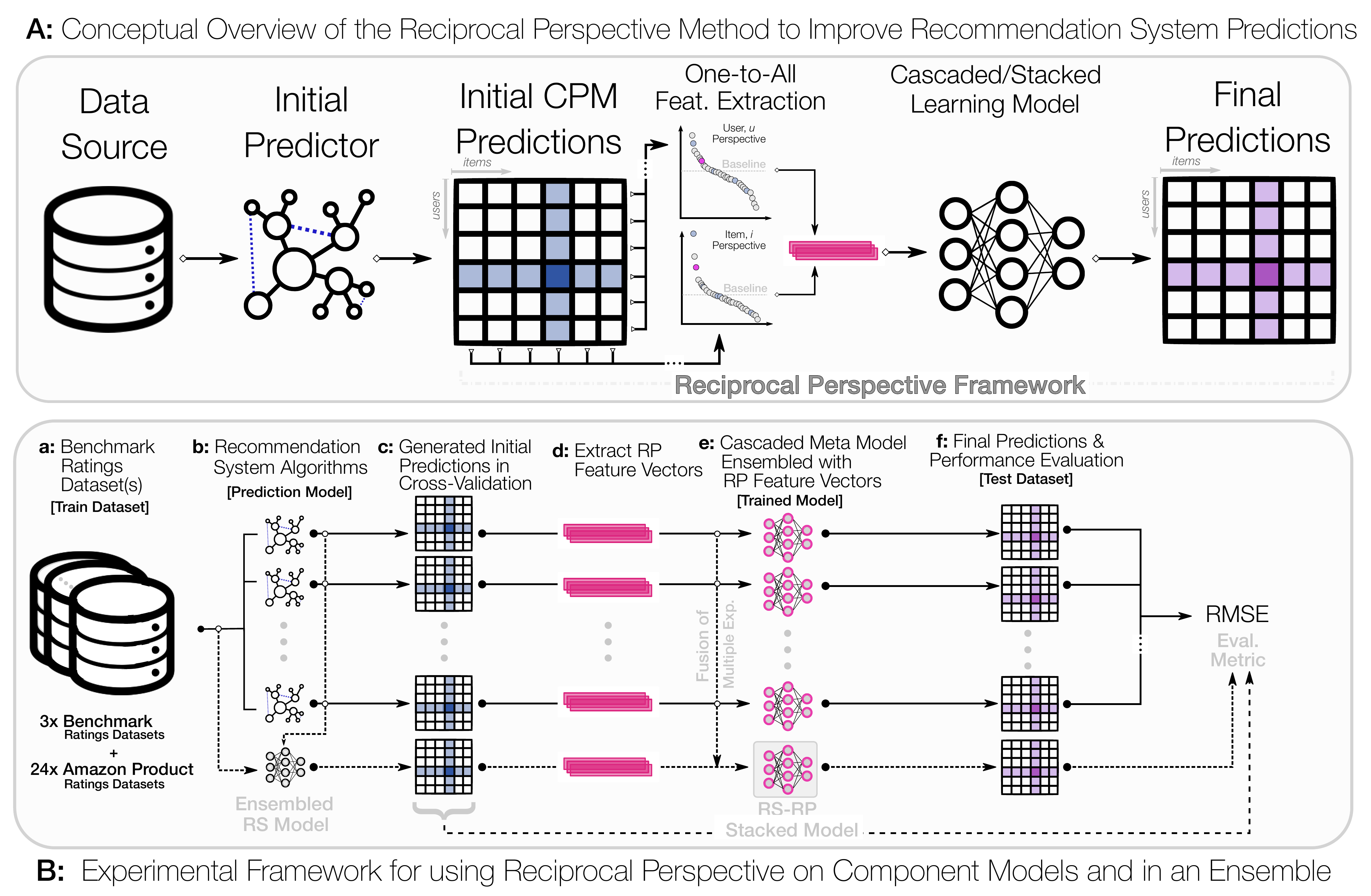}
  \caption{Conceptual Overview of the Generalized Reciprocal Perspective Framework.}
  \label{fig:overview}
\end{figure*}

Following from the brief RP methodology description above, a complete technical description of the RP framework is later described in section\ref{sec:rpgen}, however, to understand the generalization of the RP framework to a broader class of problems, we first review how RP was leveraged for each independent bioinformatic application and how these relate to generalized link prediction. A conceptual overview of the RP framework is depicted in Fig. \ref{fig:overview}A.

In each bioinformatic application domain, SOTA predictors (referred to as the \textit{initial predictors}) were leveraged to generate pairwise predictions for all possible combinations between elements. As previously described, the inference of all pairwise prediction scores between all elements forms the CPM, comprising a complete adjacency matrix representing a $K_n$ complete graph with weighted edges. The predictor-specific CPMs are then leveraged to extract pairwise RP features for use in the training the cascaded predictor. 
In the intra-species PPI tasks, we generated a square CPM, $\mathbb{R}^{n \times n}$, for all possible pairs between $n$ proteins within an organism (\textit{e.g.} $n = {\sim}21,000$ for \textit{Homo sapiens}). In this work, we denote this RP formulation as \textbf{\textit{Homo}-RP}, indicating that CPM row and column indices represent the same node type (element sets $A=B$), the CPM diagonal represents self-predictions, and the complete graph can be expressed as $K_n$.

For other prediction problem formulations between two differing sets of elements (\textit{e.g.} inter- and cross-species PPI prediction, mRNA-target interaction prediction, or drug-target interaction prediction), with element set sizes of $n$ and $m$, respectively, the resultant CPM, $\mathbb{R}^{n \times m}$, is typically non-square ($n \neq m$). In this work, we denote this RP formulation as \textbf{\textit{Hetero}-RP}, indicating that CPM row and column indices are different and distinct sets, the CPM is a complete bipartite adjacency matrix, and the complete graph can be expressed as $K_{n,m}$.

The ability to computationally generate CPMs has only recently become possible with the advent of high-performance computing infrastructure and algorithmic optimizations. While domain-specific SOTA methods are methodologically diverse, fortunately the CPM generation process is \textit{embarrassingly parallel} and computationally efficient methods can be scaled to generate such large matrices. Evidently, these matrices grow with the square of the number of elements (\textit{i.e.} $n^2$ or $nm$). In generalizing RP to the broader class of Network Science methods, it is therefore important to select a class of link predictors applicable to diverse datasets and problem formulations. To that end, we opted to demonstrate the efficacy of RP on predicting the five-point star user-item rating with diverse Recommendation System (RS) algorithms leveraged as the initial predictor. Given the previous success in leveraging RP in diverse biomedical domains with a wide variety of problem formulations and its demonstrated utility in combination with RS algorithms in this work, we seek to demonstrate the generalized utility of the RP framework for many pairwise applications.

\subsection{Reciprocal Perspective for Recommendation Systems}

The prediction of user-item relationships has considerable (eCommerce) financial consequence within diverse industries. Most notably, the study and development of RS algorithms has been greatly popularized and adopted by industry in the last 15 years. RS algorithms were widely adopted in the technology industry and active applied research reported their utility to an increasingly broad array of prediction tasks. Fortunately, RS algorithms are usually computationally efficient and scalable with sufficient supporting compute infrastructure. This class of link prediction algorithms is well-suited to evaluating the impact of applying RP to diverse application domains. To demonstrate the universal application of RP to network-based abstractions of real-world problems, we consider three different data sources and 13 different RS algorithms. We further demonstrate how RP can be leveraged as a \textit{combination of multiple experts} ensemble technique for $n \gg 2$ component models (Fig. \ref{fig:overview}B).

\section{Related Work}

The RP framework is a cascaded \textbf{semi-supervised machine learning} layer that leverages features derived from the CPM in a space unique from the original application domain \cite{dick2018reciprocal}. There exist many semi-supervised approaches for link prediction tasks \cite{kashima2009link, shahreza2017heter, berton2015link, ceci2015semi} including link propagation methods, label propagation, active learning, and multi-view learning; however, each of these approaches operate on the graph characteristics and/or feature spaces of the original domain to discover new links among disjoint nodes. The RP framework differs in that it leverages context-based features derived from the predicted scores (produced from the application of an initial predictor) of the CPM within a cascaded machine learning method. The RP framework is, therefore, analogous to the use of stacked classifiers in pattern classification/regression, however goes beyond these approaches to consider \textit{all} predictions involving either element in the pair. These derived features reside in a space (\textit{i.e.} domain agnostic) that differs from those leveraged by the initial predictor thereby differing conceptually from the other forms of semi-supervised machine learning.

RP is conceptually most similar to \textit{transductive learning} in that the few labelled training samples are situated in the context of the unlabelled data. Transductive learning algorithms learn the latent structure of the data by clustering all (un)labeled samples and then labeling a given cluster using those few samples for which labels are available, while additionally accounting for the data distribution \cite{Shi2018ECCV, Song2018CVPR}. These transduced labels are then used for prediction decisions, as exemplified in \cite{yones2018genome}. The context-based features leveraged by RP are, however, derived from predicted scores which reside in a feature-space that differs from the original feature space where transductive learning would perform its clustering.

Another fundamental facet of RP (as emphasized by its nomenclature) is \textbf{reciprocity}. The concept of reciprocity is common to Network Science applications and integrated within many methods given the intrinsic duality of pairwise relationships. Ali \textit{et al.} predicted reciprocal links within directed citation networks, where author $a_i$ creates a direct link (through citation) to disconnected author $a_j$, creating a one-way (parasocial) relationship with $a_j$ that forms a two-way (reciprocal) relationship when author $a_j$ creates a link (through citation) to $a_i$ \cite{daud2017will}. Reciprocal links in directed graphs have also been leveraged to provide additional information on directed closure triads in the work of Li \textit{et al.} \cite{li2020link}. RS algorithms themselves will incorporate modeling terms integrating factors from both set contributions as seen in matrix factorization methods such as Singular Value Decomposition (SVD) \cite{koren2009matrix}. While the reciprocity underpinning the RP framework similarly seeks to leverage information contributed from two differing views, the RP features are derived from the  pair of O2A curves sliced from the CPM. This enables the contextualization of a given pair's inferred score in the global context of (un)labelled score distributions from these reciprocal views. This differs importantly from the locally propagated factors described in the cited work that will not make use of a complete graph of inferred scores.

The final fundamental facet of the RP method is its application in a cascade for \textbf{refinement of initial predictions}. The class of calibration techniques is one that seeks to map raw prediction scores to posterior probabilities based on observed incidence rates at different score thresholds. As reported in the work of \cite{niculescu2005predicting}, the two methods for calibrating these initial model predictions are Platt Calibration and Isotonic Regression. The former transforms initial model predicitons into a posterior probability using a sigmoid transform \cite{platt1999probabilistic}, whereas the latter learns an isotonic function to adjust the calibrate (\textit{i.e.} refines) the model predictions \cite{zadrozny2002transforming}. In both cases, a held-out calibration set is required to successfully generate these prosterior probabilited \cite{niculescu2005predicting}. The RP framework differs from these methods in that the model refinement is learned in a cascaded model leveraging context-base features derived from the O2A curves.


\begin{figure*}[tb]
  \centering
  \includegraphics[width=0.95\textwidth]{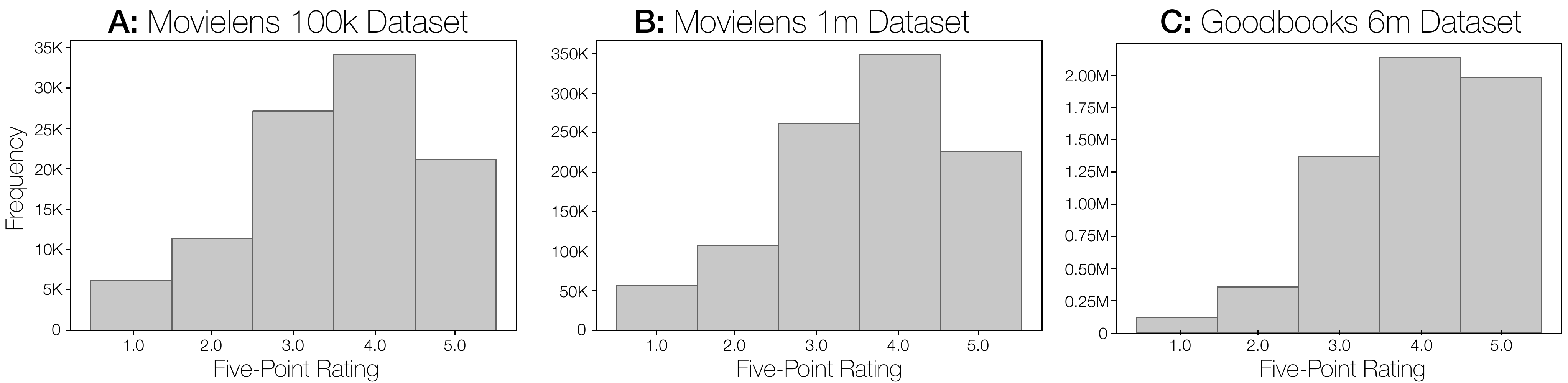}
  \caption{Distribution of Five-Point Ratings across Benchmark Datasets.}
  \label{fig:rating-dist}
\end{figure*}

\noindent \textbf{Contributions.} In this work, we generalize the RP framework outside of the bioinformatics domain to demonstrate its applicability to link prediction and machine learning problems capable of being abstracted to a network-based representation. In this work we consider the class of RS algorithms to demonstrate RP's applicability to regression-type tasks. We leverage three disparate benchmark datasets with user-item ratings on a five-point scale and apply the RP framework in a cascade to over 12 unique and scalable RS algorithms integrating various information to inform their predictions. The demonstration of RP's applicability to a regression problem here, taken together with our previous demonstration that RP is applicable to homo- and hetero-classification tasks in diverse applications within bioinforamtics, collectively support the assertion that RP is near-universally applicable to link prediction problems.

\begin{table*}[tb]
\centering
\caption{Benchmark Five-Point Ratings Scale Datasets.}
\label{tab:threedatasets}
\begin{tabular}{lccccccc}
\hline
\textbf{Dataset Source} & \textbf{\begin{tabular}[c]{@{}c@{}}Dataset\\ Alias\end{tabular}} & \textbf{\begin{tabular}[c]{@{}c@{}}Num.\\ Users\end{tabular}} & \textbf{\begin{tabular}[c]{@{}c@{}}Num. \\ Items\end{tabular}} & \textbf{\begin{tabular}[c]{@{}c@{}}Num.\\ Ratings\end{tabular}} & \textbf{\begin{tabular}[c]{@{}c@{}}CPM\\ Elements\end{tabular}} & \textbf{Density} & \textbf{Sparcity} \\ \hline
\begin{tabular}[c]{@{}l@{}}MovieLens 100K \\ Benchmark\end{tabular} & ml-100k  & 943 & 1,682  & 100,000 & 793,063 & 0.1261 & 0.8739  \\
\begin{tabular}[c]{@{}l@{}}MovieLens 1M\\ Benchmark\end{tabular}  & ml-1m  & 6,040 & 3,900  & 1,000,209 & 11,778,000 & 0.0849 & 0.9151  \\
\begin{tabular}[c]{@{}l@{}}GoodBooks 6M\\ Benchmark\end{tabular}  & gb-6m  & 53,000  & 10,000 & 6,000,000 & 265,000,000  & 0.0226 & 0.9774  \\ \hline
\end{tabular}
\end{table*}

\section{Data \& Experimental Methodology}

The following section describes the datasets considered within this study (section \ref{sec:data}), then introduces the generalized RP framework for Network Science applications (section \ref{sec:rpgen}), describes the RS algorithms that were used to produce CPMs (section \ref{sec:rsalgo}), and finally outlines all of the experimental design elements to report on the RP contribution to each of these prediction tasks.

\subsection{Rating Benchmark Datasets} \label{sec:data}

The datasets considered in this work each represent user-item ratings on a five point star rating scale and are evaluated as a regression-type prediction task. We considered the MovieLens benchmark dataset with user-movie review ratings datasets, the GoodBooks user-book review ratings dataset, and finally the Amazon Product Review datasets with a broad diversity of user-product review ratings over numerous product categories. Dataset specifics are discussed below and the distribution of five-point ratings is illustrated in Fig. \ref{fig:rating-dist}.

\subsubsection{MovieLens Benchmark Dataset} 

The MovieLens datasets are a movie RS benchmark dataset prepared by the GroupLens research group at the University of Minnesota \cite{harper2015movielens}. By collecting user-ratings from movie viewings, a user-item ratings matrix was created for the purpose of recommending movies for users based on shared interest. The MovieLens benchmarks are organized into various-sized variants according to total rating number among different movie and user-bases. In this work, we considered the MovieLens 100K benchmark (with 100K ratings distributed among 943 users and 1,682 movies) as well as the superset MovieLens 1M benchmark (with 1,000,209 ratings distributed among 6,040 users and 3,900 movies). A tabulation of the benchmark statistics including density and sparcity metrics are listed in Table \ref{tab:threedatasets}.

\subsubsection{Goodbooks Benchmark Dataset}

The GoodBooks datasets is a book RS benchmark dataset organized from the Goodreads user-book ratings online forum. The dataset has various version of differing sizes and complexities that have been used to benchmark RS algorithms as well as form the basis of numerous Kaggle competitions \cite{vinh2020hyperml, sun2022neural, slokom2018comparing}. In this work, we leveraged one of the largest versions with six million ratings between 53,000 users and 10,000 books. A tabulation of the benchmark statistics including density and sparcity metrics are listed in Table \ref{tab:threedatasets}.

\subsubsection{Amazon Review Dataset}

The Amazon Review Datasets (2018 update) was originally assembled at the University of California San Diego \cite{he2016ups, mcauley2015image}. The updated dataset contains a total of 233.1 million reviews spanning May 1996 - Oct 2018 and 24 different categories of items. When subdivided according to product category, each dataset represents a gargantuan CPM size that is incredibly sparse. We tabulate the CPM size (in number of unique predictions, and requisite RAM/storage space to manipulate the object) as well as the matrix sparcity and density in the Supplementary Materials. With a single prediction represented by a single precision float in 4 bytes, we computed CPM storage as \nicefrac{nm}{2} $\times$ 4 bytes, revealing the range of CPM storage infrastructure to be between 678 GB for the smallest and 722.6 TB for the largest.

To account for these overly sparse and intractably large CPMs, we considered the $k$-core of each product category where each user and item must provide or receive at least $k$ ratings. This space reduction approach leverages a constant thresholding across all product categories, however when ranked by number of CPM elements, there are two orders of magnitude between the smallest and largest datasets suggestive that a constant $k$ thresholding may be too extreme for the smaller datasets (too many users and items are truncated) and perhaps insufficient for the largest datasets (too few user and items are truncated). In this work, we considered a constant $k=20$ threshold to the Amazon Review datasets (see Supplementary Materials), however we also propose that a variable $k$-core should be extracted in accordance to another metric. We chose the density index and enforced a constant 1\% density index across each category (Table \ref{tab:amz1p}). This was achieved by iteratively incrementing threshold $k$ until the resultant matrix density surpassed the minimum 1\% requirement. This \textit{densification} resulted in considerably more balanced datasets and the Supplementary Materials illustrate the trade-offs in retained users, items, and ratings for each variable $k$-core threshold and the impact on the density and sparcity index. 

\begin{table*}[]
\centering
\caption{Amazon Product Rating Dataset Densification by Minimum Density}
\label{tab:amz1p}
\begin{tabular}{lccccc
>{\columncolor[HTML]{EFEFEF}}c c}
\toprule
\textbf{Product Category} & \textbf{Threshold} & \textbf{\begin{tabular}[c]{@{}c@{}}Remaining \\ Users\end{tabular}} & \textbf{\begin{tabular}[c]{@{}c@{}}Remaining\\ Items\end{tabular}} & \textbf{\begin{tabular}[c]{@{}c@{}}Remaining\\ Ratings\end{tabular}} & \textbf{\begin{tabular}[c]{@{}c@{}}CPM\\ Elements\end{tabular}} & \textbf{Density} & \textbf{Sparcity} \\ \midrule
\multicolumn{1}{r|}{Amazon Instant Video} & 9 & 1,658 & 575 & 12,028 & 476,675 & 0.0126 & 0.9874 \\
\multicolumn{1}{r|}{Apps for Android}  & 20  & 4,561 & 1,831 & 89,961 & 4,175,595 & 0.0108 & 0.9892 \\
\multicolumn{1}{r|}{Automotive}  & 14  & 2,058 & 211 & 4,777 & 217,119 & 0.0110 & 0.9890 \\
\multicolumn{1}{r|}{Baby}  & 14  & 2,829 & 896 & 25,975 & 1,267,392 & 0.010 & 0.9898 \\
\multicolumn{1}{r|}{Beauty}  & 17  & 3,529 & 887 & 32,323 & 1,565,111 & 0.010 & 0.9897 \\
\multicolumn{1}{r|}{Book}  & 83  & 12,804 & 1,739 & 228,642 & 11,133,078  & 0.010 & 0.9897 \\
\multicolumn{1}{r|}{CDs and Vinyl} & 35  & 5,932 & 1,771 & 105,590 & 5,252,786 & 0.010 & 0.9899 \\
\multicolumn{1}{r|}{Cell Phones and Accessories}  & 15  & 2,403 & 435 & 10,874 & 522,652 & 0.010 & 0.9896 \\
\multicolumn{1}{r|}{Clothing Shoes and Jewelry} & 21  & 2,296 & 120 & 3,501 & 137,760 & 0.0127 & 0.9873 \\
\multicolumn{1}{r|}{Digital Music} & 13  & 2,551 & 1,021 & 26,444 & 1,302,285 & 0.010 & 0.9898 \\
\multicolumn{1}{r|}{Electronics} & 30  & 5,176 & 919 & 51,702 & 2,378,372 & 0.0109 & 0.9891 \\
\multicolumn{1}{r|}{Grocery and Gourmet Food}  & 14  & 4,629 & 1,075 & 50,304 & 2,488,087 & 0.0101 & 0.9899 \\
\multicolumn{1}{r|}{Health and Personal Care}  & 18  & 4,216 & 1,285 & 55,785 & 2,708,780 & 0.0103 & 0.9897 \\
\multicolumn{1}{r|}{Home and Kitchen}  & 23  & 3,235 & 513 & 18,794 & 829,777 & 0.0113 & 0.9887 \\
\multicolumn{1}{r|}{Kindle Store} & 35  & 4,802 & 840 & 43,007 & 2,016,840 & 0.0107 & 0.9893 \\
\multicolumn{1}{r|}{Movies and TV} & 33  & 7,560 & 4,617 & 351,551 & 17,452,260  & 0.0101 & 0.9899 \\
\multicolumn{1}{r|}{Musical Instruments}  & 10  & 1,679 & 323 & 5,827 & 271,158 & 0.0107 & 0.9893 \\
\multicolumn{1}{r|}{Office Products}  & 11  & 2,812 & 911 & 31,988 & 1,280,866 & 0.0125 & 0.9875 \\
\multicolumn{1}{r|}{Patio Lawn and Garden} & 11  & 1,397 & 298 & 5,981 & 208,153 & 0.0144 & 0.9856 \\
\multicolumn{1}{r|}{Pet Supplies} & 15  & 2,648 & 605 & 17,299 & 801,020 & 0.0108 & 0.9892 \\
\multicolumn{1}{r|}{Sports and Outdoor} & 20  & 3,145 & 427 & 14,394 & 671,457 & 0.0107 & 0.9893 \\
\multicolumn{1}{r|}{Tools and Home Improvement} & 17  & 2,400 & 281 & 7,081 & 337,200 & 0.0105 & 0.9895 \\
\multicolumn{1}{r|}{Toys and Games} & 18  & 2,492 & 644 & 16,951 & 802,424 & 0.0106 & 0.9894 \\
\multicolumn{1}{r|}{Video Games}  & 16  & 3,037 & 1,232 & 38,185 & 1,870,792 & 0.0102 & 0.9898 \\ \bottomrule
\end{tabular}
\end{table*}

\subsection{Generalized Reciprocal Perspective Formulation}\label{sec:rpgen}

Following from the successful application of the RP framework in the bioinformatic sphere, we sought to generalize the methodology for application to any link prediction problem abstactable to a network representation and which adhere to the assumptions required of the RP framework: 
\begin{itemize}
    \item The RP framework requires that a CPM be generated using an initial predictor; that is, the problem is computationally tractable.
    \item The generation of a complete (bipartite) graph for this problem is sensible; that is, any node can be reasonably connected to any other.
    \item The One-to-All score curves generally exhibit an approximately continuous "S"- or "L"-shaped score distribution; that is, the O2A scores are not binarized or discrete/stepwise for which no reasonable baseline can be determined.
\end{itemize}
These limitations are further discussed in section \ref{sec:resultsdiscussion}. From this CPM $ = \mathbb{R}^{n \times m}$, a given user row $u$ and a given item column $i$ will specify a cell containing the initial predictor score for pair $ui$. We require that some proportion of the CPM have ground-truth values in order to train the cascaded model. It is not necessary that each user $u$ or item $i$ to have ground-truth labels given that the RP features derived from the CPM are obtained from the scores inferred by the initial predictor. 

\begin{figure}
    \centering
    \includegraphics[width=0.45\textwidth]{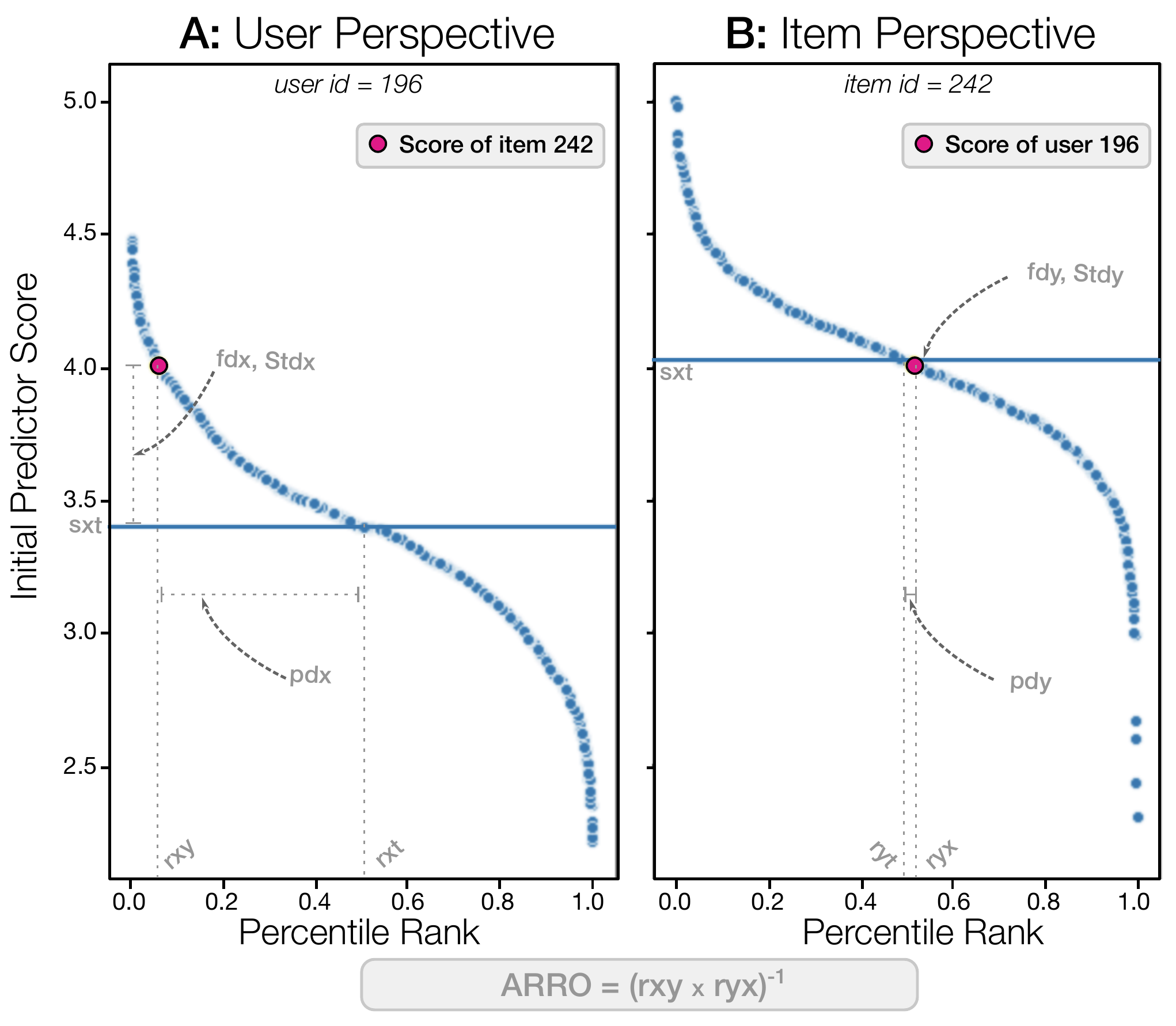}
    \caption{Example One-to-All Curves for a User-Item Pair Illustrating the 14 Derived RP Features from this Rank-Order Prediction Space. Six of the extracted RP features from each O2A curve are depicted and correspond to the definitions listed in Table \ref{tab:features}.}
    \label{fig:example-o2a}
\end{figure}

By extracting all other predicted scores within the CPM row including user $u$, we can sort those scores into rank-order to generate a \textit{One-to-All} score curve that has the initial predictor's five-point score rating on the y-axis and the normalized percentile rank-order on the x-axis. We denote this the "user perspective" for $u$ (Fig. \ref{fig:overview}). Similarly, the extraction of all predicted scores of column $i$ from the CPM, we generate a complimentary rank-order O2A. By definition, only the singular predicted score beteween $u,i$ within these two reciprocal O2A curves is shared. This complimentary distribution is denoted the "item perspective" for item $i$ (Fig. \ref{fig:overview}). A visualization of these paired O2A curves and the extracted features are depicted in Fig. \ref{fig:example-o2a}.

To more formally describe the extracted RP features, we leverage set theory notation. We first define the set of $n$ users as $U = \{u_{1},u_{2},\ldots,u_{n}\}$. Similarly the set of $m$ items is defined as $I=\{i_{1},i_{2},\ldots,i_{m}\}$. We define $u_{a}i_{b}~|~u_{a} \in U, i_{b} \in I, a \in \{1,2,\ldots,n\}, b \in \{1,2,\ldots,m\}$ as a predicted score between any user $u_a$ and item $i_b$ where $C$ is the set of all possible interactions between the elements of sets $U$ and $I$:
\begin{equation}
C = \{u_{1}i_{1},u_{1}i_{2},\ldots,u_{a}i_{b},\ldots,u_{n}i_{m-1},u_{n}i_{m}\}
\end{equation}

In the case where we consider homo-RP interactions, the element set $U$ is identical to element set $I$. Furthermore, we define the score and rank order for a given relationship, $u_{a}i_{b}$. The score of a given rating, $u_{a}i_{b}$, is defined as $s_{u_{a}i_{b}}$ for the user-perspective and $s_{i_{b}u_{a}}$ for the item-perspective and these values are used to sort the respective sets. The sorted distribution of scores into monotonically decreasing rank-order enables the creation of the O2A for each user-item pair. Given the (typically) differing sizes $n \neq m$, the natural-numbered rank positions of each element are normalized into a real-numbered \textit{percentile} rank to consistently express rank-type features within the numerical range $[0,1]$. Characteristic to these curves is a low-rank/high-scoring region, a baseline region, and a high-rank/low-scoring region. In previous work, the baseline for each of the L-shaped O2As was determined using a knee-detection method that sought to identify threshold $\tau$ that delineates the point between the low-rank/high-scoring pairs and the subsequent baseline. In this work, given the S-shaped distributions, we approximate the baseline with the median score. 

A predicted score can then be appraised in the context of its relative position in each O2A curve, as well as relative to the O2A baseline. From a complete CPM, these paired O2A curves can be extracted and all 14 RP-defined features can be extracted for subsequent use in a cascaded model. A definition of the 14 RP defined features is listed in Table \ref{tab:features}. The extracted features may be classified as "score"-, "rank"-, "fold"-, or "stats"-types, depending on their definition. For example, the fold-type features seek to measure the relative distance between a given score and the O2A curve's baseline value: 
\begin{equation}
FD_{u_{a}i_{b}}~=~\frac{s_{u_{a}i_{b}}~-~s_{u_{\tau}}}{s_{u_{\tau}}}
\end{equation}
\begin{equation}
FD_{i_{b}u_{a}}~=~\frac{s_{i_{b}u_{a}}~-~s_{i_{\tau}}}{s_{i_{\tau}}}
\end{equation}

In essence, these features seek to contextualize the relative position of a given element within the reciprocal perspective according to the distribution of all other elements within that perspective (Fig. \ref{fig:example-o2a}). These extracted features can then be leveraged in the training of a cascaded machine learning model; typically an eXtreme Gradient Boosted (XGBoost) model \cite{chen2015xgboost}.

To learn a cascaded model, we do not need to extract features from every possible pair since the vast majority are unlabeled. The unlabelled initial prediction scores are used to define the morphology of the O2A curve as in transductive learning. Thus, we make use of the O2A data distributions generated from the predicted CPM in a semi-supervised framework where both labelled and unlabelled initial predictor scores inform the feature extraction and the cascaded learning algorithm learns to effectively \textit{correct} the initial predictions leading to improved task-specific predictions. We clarify that the RP features need only be computed for labelled pairs to train the cascaded model, however the derived features benefit from the distribution of the unlabelled initial predictor scores. 

Supervised learning algorithms often rely upon fully labeled datasets that are typically expensive to create. An attractive alternative are semi-supervised learning (SSL) \cite{chapelle2006semi} algorithms which relax the need for labels by learning the structure of the data from the unlabeled samples. Numerous studies have demonstrated that the achievable performance of SSLs approaches those of fully supervised learning despite requiring only fractions of the data to be labeled. Recent examples include \cite{laine2016temporal,tarvainen2017mean} wherein data from an existing classification dataset (\textit{e.g.} CIFAR-10 \cite{krizhevsky2014cifar}) are randomly sampled and the remaining examples treated as unlabeled.



In this work, we rely upon both supervised and SSL in an ensemble, context-based, machine learning pipeline. The fully-labeled user-item ratings datasets, $(\boldsymbol{x},y) \in \mathcal{D}$, are used in supervised training to generate our user-item predictor, $\hat{f_\theta}(\cdot)$. We then use $\hat{f_\theta}(c)$ to predict a score, $u_{a}i_{b}$, for each possible user-item pair generating a CPM comprising a complete bipartite graph:
\begin{equation}
   s_{u_{a}i_{b}} = \hat{f_\theta}(c)~\forall~ c \in C
\end{equation}

In this work, we seek to most accurately predict numeric user-item ratings and therefore report performance based on the root-mean squared error (RMSE) metric for each of the benchmark datasets and RS algorithms considered in this study:

\begin{equation}
     RMSE = \sqrt{\frac{1}{n}\Sigma_{i=1}^{n}{\big(\hat{y}_i -y_i\big)^2}}
\end{equation}

\begin{table*}[]
\centering
\caption{The 14 RP Features derived from Pair-Specific One-to-All Score Curves.}
\label{tab:features}
\begin{tabular}{lccl}
\toprule
\textbf{Feature Generic Name} & \textbf{Short Name}  & \textbf{Type} & \textbf{Description} \\ \midrule
Y-in-X-percentile & \textit{ryx} & Rank & Percentile of element Y among all the predictions for element X \\
X-in-Y-percentile & \textit{rxy} & Rank & Percentile of element X among all the predictions for element Y \\
Adjusted reciprocal rank order  & ARRO   & Rank & Reciprocal product of \textit{rxy} and \textit{ryx}  \\
X-percentile-baseline  & \textit{rxt} & Rank & Percentile rank of the element nearest to the local cutoff value of element X \\
X-baseline & \textit{sxt} & Score  & Score at the local cutoff value of element X \\
Y-percentile-baseline  & \textit{ryt} & Rank & Percentile rank of the element nearest to the local cutoff value of element Y \\
Y-baseline & \textit{syt} & Score  & Score at the local cutoff value of element Y \\
Percentile-difference-from- baseline-X & \textit{pdx }& Fold & Difference between \textit{rxy} and \textit{rxt} \\
Percentile-difference-from- baseline-Y & \textit{pdy} & Fold & Difference between \textit{ryx} and \textit{ryt} \\
Fold-difference-from-baseline-X & \textit{fdx} & Fold & Fold-difference of element Y score in element X from baseline \textit{sxt} \\
Fold-difference-from-baseline-Y & \textit{fdy} & Fold & Fold-difference of element X score in element Y from baseline \textit{syt} \\
SD-distance-from-mean-X  & \textit{Stdx}   & Stats  & The number of standard deviations from the mean score in element X  \\
SD-distance-from-mean-Y  &\textit{ Stdy}   & Stats  & The number of standard deviations from the mean score in element Y  \\
Original-Score & \textless{}pred-name\textgreater{} & Score  & The originally predicted score from the initial predictor   \\ \bottomrule
\end{tabular}
\end{table*}

\subsection{Recommendation Systems Algorithms} \label{sec:rsalgo}

This work makes use of twelve Recommendation System (RS) algorithms to generate independent CPMs for each of the datasets considered in this work. RS algorithms are ubiquitous in user-item rating prediction, they have typically fast inference times that can scale to produce complete CPMs, and each leverage a diversity of information. 

\begin{table}[tb]
\caption{Mathematical Notation for Recommendation System Algorithms; adapted from \cite{sksurprise}.}
\label{tab:notation}
\begin{tabular}{cp{6cm}}
\toprule
\textbf{Notation}  & \textbf{Meaning}  \\ \midrule
\multicolumn{1}{l|}{$R$}  & the set of all ratings \\
\multicolumn{1}{l|}{\begin{tabular}[c]{@{}l@{}}$R_{train}$\\$R_{test}$\\$\hat{R}$\end{tabular}} & denote the training set, the test set, and the set of predicted ratings. \\
\multicolumn{1}{l|}{$U_i$}  & the set of all users where $u$ and $v$ represent distinct users. \\
\multicolumn{1}{l|}{$U_{ij}$}  & the set of all users that have rated item $i$. \\
\multicolumn{1}{l|}{$I$}  & the set of all items where $i$ and $j$ represent distinct items. \\
\multicolumn{1}{l|}{$I_u$}  & the set of all items rated by user $u$.  \\
\multicolumn{1}{l|}{$I_{uv}$}  & the set of all items rated by both users $u$ and $v$. \\
\multicolumn{1}{l|}{$r_{ui}$}  & the ground-truth rating of user $u$ for item $i$.  \\
\multicolumn{1}{l|}{$r_{ui}$}  & the predicted/estimated rating for user $u$ of item $i$.  \\
\multicolumn{1}{l|}{$b_{u}$} & the baseline rating of user $u$ for item $i$. \\
\multicolumn{1}{l|}{$\mu$} & the mean of all ratings.  \\
\multicolumn{1}{l|}{$\mu_u$} & the mean of all ratings given by user $u$.  \\
\multicolumn{1}{l|}{$mu_i$} & the mean of all ratings given to item $i$.  \\
\multicolumn{1}{l|}{$\sigma_u$} & the standard deviation of all ratings given by user $u$. \\
\multicolumn{1}{l|}{$\sigma_i$} & the standard deviation of all ratings given to item $i$. \\
\multicolumn{1}{l|}{$N^{k}_{i}(u)$} & the $k$ nearest neighbours of user $u$ that have rated item $i$.  \\
\multicolumn{1}{l|}{$N^{k}_{u}(i)$} & the $k$ nearest neighbours of item $i$ that are rated by user $u$.  \\ \bottomrule
\end{tabular}
\end{table}

To implement each RS algorithm, we leveraged the SciKit-Surprise library that makes accessible many of the traditional and most commonly used RS algorithms for pairwise predictions between a given user $u$ and a given item $i$ \cite{sksurprise}. In this section we briefly define the twelve RS algorithms considered in this work. The complete set of mathematical notation used to define each method is tabulated in Table \ref{tab:notation}. Many of the classical similarity-based algorithms are implemented according to Aggarwal \textit{et al.} \cite{aggarwal2016recommender} and implementation-specific references are provided where necessary.

\vspace{1mm}
\noindent\textbf{Uniform-Random Predictor.}
\noindent Beyond the twelve RS methods listed below, we defined an additional uniform random baseline method, denoted "Negative-Control", where rating "predictions" between user $u$ and item $i$, denoted $\hat{r}_{ui}$, were drawn from a uniform random distribution over the discrete set of possible five-point ratings: $\{1.0, 2.0, 3.0, 4.0, 5.0\}$. A series of negative control experiments are defined in section \ref{sec:randexp-method} to which this method and the following Random-Normal predictor contribute.

\vspace{1mm}
\noindent\textbf{Random-Normal Predictor.}
\noindent A baseline RS algorithm predicting a random rating based on the distribution of the training set (assumed to be normal). The prediction $\hat{r}_{ui}$ is generated from a $\mathcal{N}(\hat{\mu}, \hat{\sigma}^2)$ where the prediction mean $\hat{\mu}$ and standard deviation $\hat{\sigma}$ estimated from the training data using Maximum Likelihood Estimation:

\begin{equation}
  \begin{split}
 \hat{\mu} &= \frac{1}{|R_{train}|} \sum_{r_{ui} \in R_{train}}r_{ui}\\\\   
 \hat{\sigma} &= \sqrt{\sum_{r_{ui} \in R_{train}}
 \frac{(r_{ui} - \hat{\mu})^2}{|R_{train}|}}
  \end{split}
\end{equation}

\vspace{1mm}
\noindent\textbf{Baseline Predictor.}
\noindent An RS algorithm predicting the baseline estimate for given user and item. The baseline is considered the global mean rating, modulated by bias terms:

\begin{equation}
  \hat{r}_{ui} = b_{ui} = \mu + b_u + b_i
\end{equation}

\noindent If user $u$ or item $i$ is unknown, then the biases $b_u = 0$ and/or $b_i = 0$, respectively.

\vspace{1mm}
\noindent\textbf{$k$-Nearest Neighbours Basic (KNN).}
\noindent The basic KNN predictor implements a collaborative filtering algorithm that generates prediction $\hat{r}_{ui}$ depending on whether element similarities are measured between users or between items, which is reported to have a considerable impact on the results. For example, the prediction $\hat{r}_{ui}$ for user-based similarity is given as:

\begin{equation}
  \hat{r}_{ui} = \frac{\sum\limits_{v \in N^k_i(u)} \text{sim}(u, v) \cdot r_{vi}}
  {\sum\limits_{v \in N^k_i(u)} \text{sim}(u, v)}
\end{equation}

\noindent and the complementary item-specific formulation by: 

\begin{equation}
    \hat{r}_{ui} = \frac{\sum\limits_{j \in N^k_u(i)} \text{sim}(i, j) \cdot r_{uj}}{\sum\limits_{j \in N^k_u(i)} \text{sim}(i, j)}
\end{equation}

\noindent Implementation-specific details are available at \cite{sksurprise}.

\vspace{1mm}
\noindent\textbf{$k$-Nearest Neighbours with Means (KNN-Means).}
\noindent In a slight expansion on the basic KNN algorithm, the KNN RS accounts for the mean rating of each user and is given as:

\begin{equation}
  \hat{r}_{ui} = \mu_u + \frac{ \sum\limits_{v \in N^k_i(u)} 
  \text{sim}(u, v) \cdot (r_{vi} - \mu_v)} {\sum\limits_{v \in ^k_i(u)} \text{sim}(u, v)}
\end{equation}

\noindent and the complementary item-specific formulation and related details are available at \cite{sksurprise}.

\vspace{1mm}
\noindent\textbf{$k$-Nearest Neighbours with Z-Score (KNN-Zscore).}
\noindent In a further expansion on the basic KNN algorithm, the KNN RS accounts for the Z-score rating (a quantification of the distance measured in number of standard deviations below or above the population mean a raw score is) of each user and is given as:

\begin{equation}
  \hat{r}_{ui} = \mu_u + \sigma_u \frac{ \sum\limits_{v \in N^k_i(u)} \text{sim}(u, v) \cdot (r_{vi} - \mu_v) / \sigma_v} {\sum\limits_{v \in N^k_i(u)} \text{sim}(u, v)}
\end{equation}

\noindent and the the complementary item-specific formulation and related details are available at \cite{sksurprise}.

\vspace{1mm}
\noindent\textbf{$k$-Nearest Neighbours with Baseline (KNN-Baseline).}
\noindent A basic collaborative filtering algorithm taking into account a baseline rating:

\begin{equation}
  \hat{r}_{ui} = b_{ui} + \frac{ \sum\limits_{v \in N^k_i(u)} \text{sim}(u, v) \cdot (r_{vi} - b_{vi})} {\sum\limits_{v \in N^k_i(u)} \text{sim}(u, v)}
\end{equation}

\vspace{1mm}
\noindent\textbf{Singular Value Decomposition (SVD).}
\noindent The famous SVD algorithm, as popularized by Simon Funk during the Netflix Prize. The predicted rating $\hat{r}_{ui}$ is given as:

\begin{equation}
  \hat{r}_{ui} = \mu + b_u + b_i + q_i^Tp_u
\end{equation}

\noindent To estimate all of the equation unknowns ($b_u, b_i, p_u, q_i$), we minimize the following regularized squared error and using stochastic gradient descent:

\begin{equation}
  \sum_{r_{ui} \in R_{train}} \left(r_{ui} - \hat{r}_{ui} \right)^2 + \lambda\left(b_i^2 + b_u^2 + ||q_i||^2 + ||p_u||^2\right)
\end{equation}

\vspace{1mm}
\noindent\textbf{Singular Value Decomposition with Implicit Ratings (SVD++)}
\noindent The SVD++ algorithm extends the SVD algorithm by accounting for implicit ratings. That is, the predicted rating $\hat{r}_{ui}$ also contains $y_j$ terms representing the set of item factors capturing implicit ratings which capture the fact that a user $u$ rated an item $j$ regardless of the rating value:

\begin{equation}
  \hat{r}_{ui} = \mu + b_u + b_i + q_i^T\left(p_u + |I_u|^{-\frac{1}{2}} \sum_{j \in I_u}y_j\right)
\end{equation}

\vspace{1mm}
\noindent\textbf{Non-negative Matrix Factorization (NMF).}
\noindent A collaborative filtering algorithm based on Non-negative Matrix Factorization (NMF) similar to SVD where regularized stochastic gradient descent is used as model optimization procedure \cite{luo2014efficient}. A predicted rating $\hat{r}_{ui}$ is given as:

\begin{equation}
  \hat{r}_{ui} = q_i^Tp_u
\end{equation}

\noindent where user and item factors are kept strictly positive and each SGD step updates factors according to:

\begin{equation}
  \begin{split}
 p_{uf} \leftarrow p_{uf} \cdot \frac{\sum_{i \in I_u} q_{if}
 \cdot r_{ui}}{\sum_{i \in I_u} q_{if} \cdot \hat{r_{ui}} +  \lambda_u |I_u| p_{uf}}\\
 q_{if} \leftarrow q_{if} \cdot \frac{\sum_{u \in U_i} p_{uf}
 \cdot r_{ui}}{\sum_{u \in U_i} p_{uf} \cdot \hat{r_{ui}} +  \lambda_i |U_i| q_{if}}\\
  \end{split}
\end{equation}

\noindent with user- and item-specific regularization factors $\lambda_u$ and $lambda_i$, respectively. Implementation-specific details are described in \cite{luo2014efficient}.

\vspace{1mm}
\noindent\textbf{Slope One.}
\noindent The SlopeOne algorithm is a collaborative filtering algorithm that integrates information from users that rated the same item as well as information from the other items rated by the same user. Specifics for the implementation of the SlopeOne algorithm are described in \cite{lemire2005slope} and the predicted rating $\hat{r}_{ui}$ is given as: 

\begin{equation}
  \hat{r}_{ui} = \mu_u + \frac{1}{|R_i(u)|}
  \sum\limits_{j \in R_i(u)} \text{dev}(i, j),
\end{equation}

\noindent where $R_i(u)$ represents the set of \textit{relevant items} (see \cite{lemire2005slope}), and $\text{dev}(i,j)$ (the deviance) captures the average difference between the ratings of $i$ and those of $j$; formally:

\begin{equation}
  \text{dev}(i, j) = \frac{1}{|U_{ij}|}\sum\limits_{u \in U_{ij}} r_{ui} - r_{uj}
\end{equation}

\vspace{1mm}
\noindent\textbf{Co-Clustering.}
\noindent The Co-Clustering  algorithm is a collaborative filtering algorithm that assigns users and items to respective clusters $C_u$ and $C_i$ and some to joint clusters $C_{ui}$ such that the predicted rating $\hat{r}_{ui}$ is given as: 

\begin{equation}
  \hat{r}_{ui} = \overline{C_{ui}} + (\mu_u - \overline{C_u}) + (\mu_i - \overline{C_i})
\end{equation}

The overline notation represents the average rating within that co-cluster; implementation details are further described in \cite{george2005scalable}.

\vspace{1mm}
\noindent\textbf{Ensemble.}
\noindent The Ensemble algorithm, without the application of RP, is an equi-weighted predictor combining all other 12 predictors. When applying the RP learning layer, the Ensemble method is a stacking method that fuses the RP feature vectors of each of the 12 RS CPMs that are then used to learn an XGBoost model. Rather than leverage only the 14 RP features of a single component model, we concatenate the $12 \times 14 = 168$ features into a single feature vector. While the most computationally demanding method, this model integrates the greatest amount of information to inform its predictions.

\subsection{Negative Control Experiments}\label{sec:randexp-method}

In addition to our usage of various definitions of random and baseline predictions, we sought to better understand the utility of the cascaded RP method through an additional negative control experiment where, rather than consider CPMs for randomly sampled "predictions", we instead randomized the 14 derived RP features themselves. This experiment sought to elucidate the influence of the RP cascaded learning layer when applied to fully random CPMs.

To this end, we bypassed the CPM-generation stage altogether and randomized the RP features while considering the same ground-truth training and test labels for the cascaded RP model evaluation framework. This experiment illustrated that the inclusion of an RP cascaded learning layer will, at minimum, learn to predict the mean score of the input distribution. That is, even for a uniform-random CPM, if the ground-truth labels are normally distributed, the cascade RP layer will learn to predict the mean ground-truth rating. To fully assess the performance improvement attributable to the RP framework, a cross-validation framework is required.

\subsection{Cross-Validation Experimental Framework}

The evaluation of the RP-supplied contribution in prediction performance was performed independently upon each dataset and using an independent RS algorithm. For each of the benchmark datasets considered in this work, we consistently partitioned the dataset in six randomized and eqi-sized folds. The same pre-defined folds from each dataset were used to evaluate each of our models.

The sixth fold was withheld from any training and validation work and reserved to assess the performance of the resultant models. The remaining equi-sized 5-folds were then leveraged as part of a five-fold cross-validation regime. That is, for each dataset and specific RS algorithm, the learning algorithm leveraged four of the five folds for training data and was intermediately evaluated. Each of the RS algorithms for a given dataset were provided the same access to learnable information (both as an RS-alone, and with RP). This approach enabled fair comparisons between all algorithms.

\subsection{Significance Testing}

Finally, to evaluate the statistical significance in applying RP in a cascade of the original recommendation system algorithm, we used the the Wilcoxon Signed-Rank test \cite{wilcoxon1992individual}, a non-parametric test of the null hypothesis that the distribution underlying the RS-only distribution is the same as the distribution underlying the RS+RP distribution (effectively a means to test the difference in location between distributions) with matched samples. Each test of significance was performed on a per-algorithm basis between the non-RP and RP conditions.

\subsection{Compute Infrastructure}

All experiments were performed using Compute Canada high-performance computing infrastructure. Where possible, a consistent set of compute resources were provided. For CPM generation of a given dataset with a given algorithm (including the Amazon product rating dataset at 1\% density index), each compute session was allotted 64 GB of RAM, 4 CPUs, and 12 hours of run-time. The $k=20$ core CPM generations. The Goodbooks benchmark dataset represented the largest non-densified CPM and was allotted 128 GB of RAM. Given computing constraints, 5/13 algorithms were excluded from the gb-6m analyses. As previously discussed, an important assumption of the RP framework is that it cannot be applied to problems for which the CPM cannot be computed; further limitations are described in section \ref{sec:limits}.

\section{Results \& Discussion}\label{sec:resultsdiscussion}

The Reciprocal Perspective framework has been successfully applied in three independent bioinformatic prediction tasks: PPI prediction, miRNA-target prediction, and drug-target interaction prediction. This work sought to generalize the RP framework for use in the much broader classes of problems within Network Science: recommender systems. To that end, we considered three sources for benchmark ratings data and a diverse collection of RS algorithms to demonstrate the utility of RP in applications outside of the field of bioinformatics.

\subsection{RP Significantly Improves Link Prediction Performance}
To summarize the performance improvement attributable to the cascaded application of RP, we tabulated for each dataset and RS algorithm the RMSE performance improvement. The base performance is established from the application of of the RS algorithm alone and the difference in performance of the RS+RP cascade is listed. Negative values indicate better performance while positive values indicate worse performance. Additionally, we coloured each cell based on the following statistical significance criteria:
\begin{itemize}
    \item[\textcolor{sig-green}{$\blacksquare$}] RP significantly better than non-RP ($p < 0.05$).
    \item[\textcolor{sig-blue}{$\blacksquare$}] RP better, but not significantly.
    \item[\textcolor{sig-yellow}{$\blacksquare$}] RP worse, but not significantly.
    \item[\textcolor{sig-red}{$\blacksquare$}] RP significantly worse than non-RP ($p < 0.05$).
\end{itemize}

Table \ref{tab:result-mlgb} summarizes the results on the MovieLens and GoodBooks datasets using each of the 13 RS algorithms both with and without RP whereas Table \ref{tab:result-1PDensities} summarizes results across each of the Amazon Product Category datasets.

Excitingly, across the majority (221/272) of the experiments over the three benchmark dataset sources and over the broadly diverse dataset product categories, the application of the cascaded RP layer statistically significantly ($p > 0.05$) improved the performance over the RS-alone application; significance was measured by the Wilcoxon signed rank test. In Table \ref{tab:result-mlgb}, only a single instance (Baseline RS applied to the MovieLens 100K dataset) did we not see significant improvement. The results from the Amazon Product category datasets are more varied with entire datasets (\textit{e.g.} Automotive, Musical Instruments) showing that RP actually results in significantly worse performance than the RS algorithm. However when interpreting these results across all recommendation systems, it is clear that no singular method performs well with a mixture of results. The poor performance may therefore be attributable to the an intrinsically challenging nature of these two datasets. In removing these two rows from the tabulated results, we note that only the Ensemble method contains significantly worse performance when applying RP. The equi-weighted Ensemble RS method combines the predictions of all other 11 RS algorithms to infer its predictions and this is consistently the best performing model across all datasets (see Supplementary Materials for a full tabulation). Thus, the application of RP on top of this ensemble is not expected to consistently achieve significantly improved results. These results corroborate a similar finding from the previous application of RP for miRNA-mRNA target prediction where both component model ensembles and single-model RP cascaded models would achieve a similar level of performance \cite{kyrollos2020rpmirdip}. Consequently, we can expect that the stacking of 11 models with RP would have mixed results when compared to an ensemble of those 11 over all datasets. Finally, if we were to further remove the ensembled column from Table \ref{tab:result-1PDensities}, the remaining experiments would indicate that in no case does RP result in significantly worse performance. Promisingly, for some datasets we observe a reduction of as much as 0.1 in RMSE due to RP (\textit{e.g. ml-100k}) whereas the more challenging Amazon datasets indicate smaller margins of improvement.

\subsection{RP Ensemble Successfully Fuses Combinations of Multiple Experts ($n \gg 2$)}

Over the MovieLens and Goodbooks datasets, the consistently most performant RS model is the "Ensemble" model, representing the fusion of all other RS models (Table \ref{tab:result-mlgb}). The model, even prior to the application of RP, demonstrates considerably improved performance (consistent with \cite{feuerverger2012statistical}) and the subsequent application of RP as a means of stacking multiple experts results in further considerably improved performance (Table \ref{tab:result-mlgb}, consistent with \cite{dick2021multi, kyrollos2020rpmirdip, musdti}). While the training of numerous RS algorithms and the inference times required to generate multiple algorithm-specific CPMs is computationally taxing, the benefits for doing so are clearly represented within our findings. Conveniently, the training of a given RS algorithm and the inference of the CPM need only be performed once. 

\subsection{Negative Control Experiments}\label{sec:randexp}

We note that the RS algorithms denoted Negative-Control, Random-Normal, and Baseline may confer the expectation that little to no improvement from the application of RP is expected; however, their interpretation is more nuanced given that not all random, baseline, or negative control methodologies result in the same expected performance. While these RS algorithms alone achieve a certain "no-skill" level of performance, the application of the cascaded RP learning layer captures some information from the distribution of the unlabelled pairs within the CPM.

\begin{figure}[tb]
  \centering
  \includegraphics[width=0.5\textwidth]{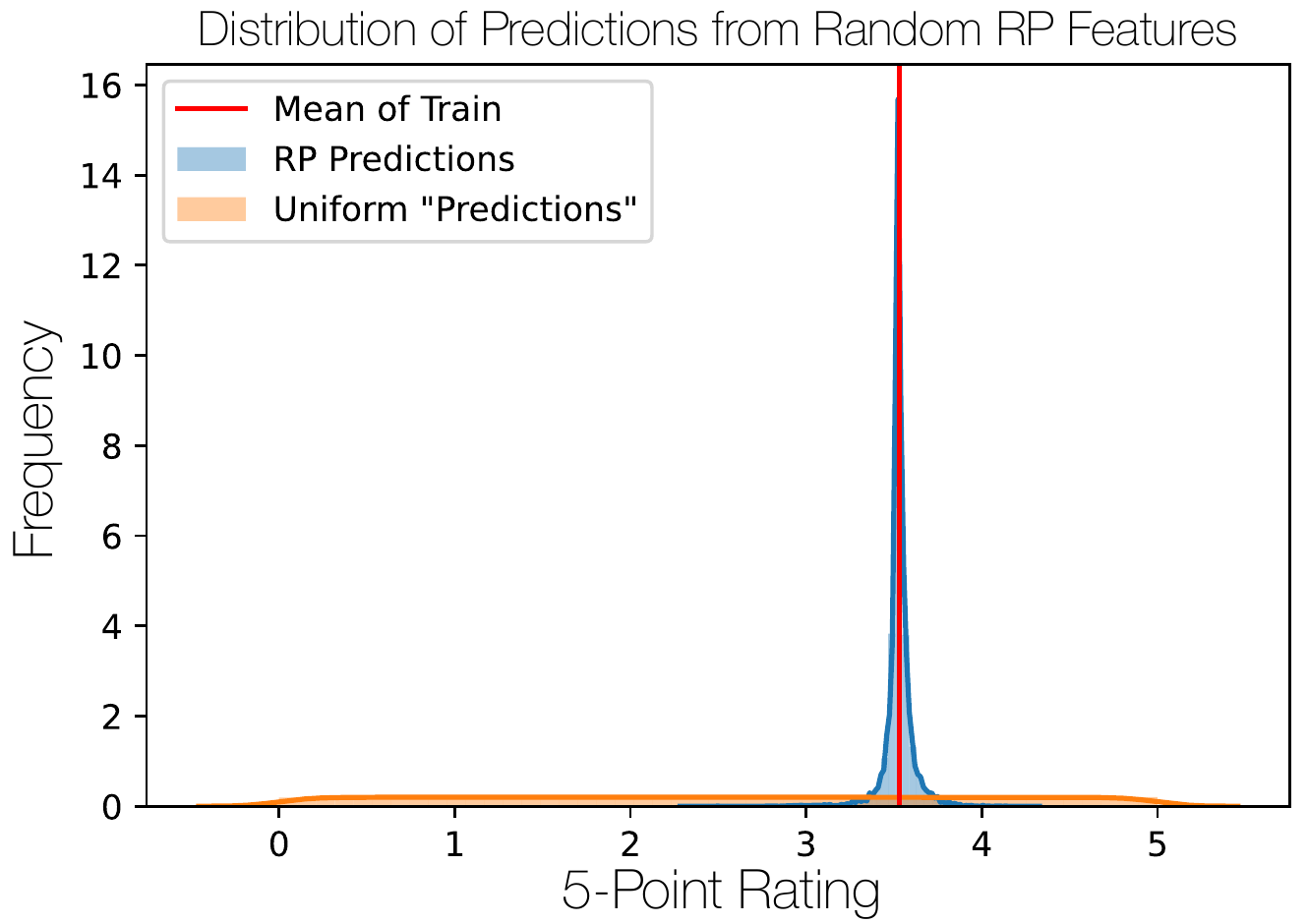}
  \caption{Distribution of XGBoost Predictions when Trained on Randomly Generate RP Features. We generate random RP features for each of the training samples of the ml-100k dataset and visualized the distribution of the resulting random model predictions when applied to the evaluation dataset. We observe that the model effectively predicts the mean value represented in the training dataset (represented in red).}
  \label{fig:randomfeat}
\end{figure}

To determine the impact of the RP learning layer application to a randomized CPM as compared to a fully randomly generated RP feature set, we generated completely random features and then trained an XGBoost model. This experiment, when evaluated on the random CPM, resulted in the XGBoost model converging its predictions on the \textit{mean rating}, thereby effectively "improving" the MSE of the initial "random" model (Fig. \ref{fig:randomfeat}).

We confirm that there is no data leakage explaining the improved performance since all models would have benefited from such leakage and several clearly do not; the Supplementary Materials depict several instances where the "Baseline" RS predictor (that generates the mean rating of a given user and item) does not benefit from the application of RP. These experiments confirm that the utility of applying RP comes from the contextual information extracted from the CPM in the SSL framework.


\subsection{Limitations on RP Applications}\label{sec:limits}

The RP framework is founded upon a number of assumptions that may limit its applicability to certain problems. We discuss a number of these below to guide potential users and propose potential solutions to reformulate a given problem into one amenable to leveraging RP.

\textbf{Computational Tractability of the CPM. }
The RP framework assumes that an initial predictor can be scaled in such a way that the problem-specific CPM may be generated. Given that certain problem domains tend to grow exponentially (\textit{e.g.} Amazon user-base and product offerings), this may make certain applications computationally intractable. Future work will investigate whether a fractional random subsets of the CPM would be sufficient to trade-off computational investment with improved predictive performance.

\textbf{Possibility for Complete Graph Connections. }
The RP framework, as exemplified in all its prior domain-specific application tasks, implicitly assumes that a connection between all nodes of a complete (bipartite) graph are indeed possible and sensible. Certainly, problem-specific constraints may violate this assumption. For example, in route navigation, certain geolocation nodes are only accessible by navigating through intermediary nodes along a path; link prediction between unconnected nodes doesn't adhere to these problem-specific constraints and therefore generating a CPM is not sensible. In such a case, one approach is to artificially "mask" out these instances by incorporating a binary-type feature indicating "(im)possible connection". Alternatively, we might reformulate the original problem into one adjusted to complete graphs (\textit{e.g.} merging unconnectable nodes into "meta-nodes" that are fully connectable). Future work may consider adapting a variant of RP to account for incomplete CPMs and/or as an interconnection proposal method.

\textbf{A Smooth-like Decay in Initial Predictor Rank-Order Predictions. }
Initial predictors producing step-wise outputs result in O2A curves that may not be amenable to RP feature extraction. In extreme step-wise contexts, the point of greatest rate of change (that we would intuitively assume to be the knee delineating high-scoring elements from the baseline in an L-shaped O2A curve) is often identified at the first point of "discontinuity" where the curve transitions from the first step to the next. Moreover, the meaningfulness of a pair's percentile rank-order is lost when this subset of equal-scoring pairs are shared in the same collective "step" (the first and last pair of a common step have the same score but vastly differing ranks). Some of these concerns are explored in \cite{dick2018fitting} and potential solutions may seek to either transform the output of the initial predictor and/or incorporate additional information that may further delineate the resultant data points.


\begin{table*}[]
  \centering
  \caption{Summary of Recommendation System RMSE Performance over Three Ratings Benchmark Datasets. Each cell represents the difference in RMSE between the original RS algorithm predictions and then the subsequent application of RP. Cell-colouring indicates; green: RP significantly better than non-RP, blue: RP better but not significantly, yellow: RP worse but not significantly, red: RP significantly worse.}
  \includegraphics[width=\textwidth]{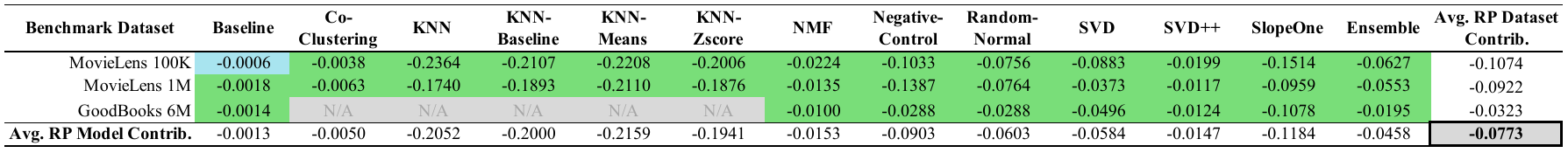}
  \label{tab:result-mlgb}
\end{table*}

\begin{table*}[]
  \centering
  \caption{Summary of Recommendation System RMSE Performance on each Amazon Product Sub-Category from Applying RP over the Consistent 1\% Matrix Densification Experiment. Each cell represents the difference in RMSE between the original RS algorithm predictions and then the subsequent application of RP. Cell-colouring indicates; green: RP significantly better than non-RP, blue: RP better but not significantly, yellow: RP worse but not significantly, red: RP significantly worse.
  }
  \label{tab:result-1PDensities}
  \includegraphics[width=\textwidth]{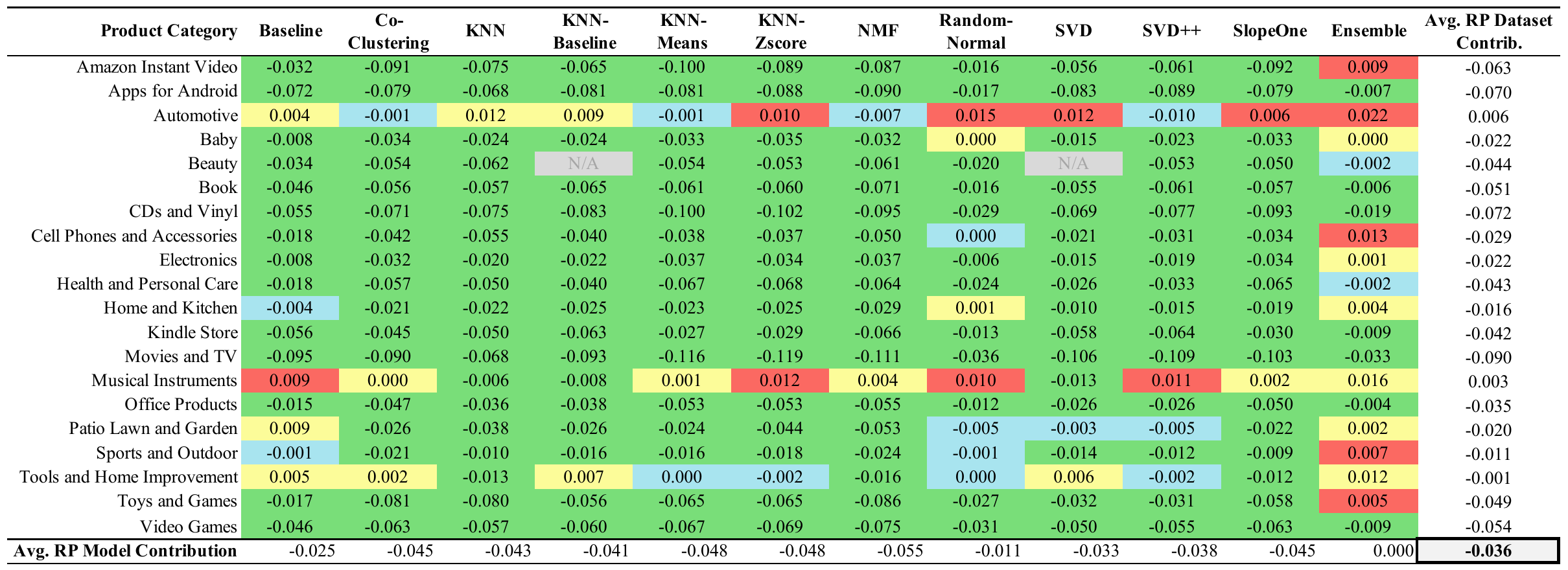}
\end{table*}

\section{Conclusion}

Real-world problems that can be reframed as a link prediction task benefit from generally applicable methods within the field of Network Science. In this work, we generalize for all pairwise link prediction tasks our novel semi-supervised machine learning method, denoted Reciprocal Perspective (RP), which refines pairwise prediction scores by placing them in the context of all scores for all potential links involving either element of the pair. Context-based features are extracted from the CPM for use in a cascaded model which has demonstrated significant improvement in predictive performance. We leveraged three independent data sources and 13 RS algorithms to demonstrate that the application of RP in a cascade almost always results in significantly ($p < 0.05$) improved predictions. These regression results, taken together with previous demonstrations of RP's near universal benefit for classification tasks, lead us to recommend RP for (m)any pairwise link prediction pipelines for which the CPM may be generated. The RP framework is available for use from the following GitHub repository: \url{https://github.com/GreenCUBIC/RP}.

\section*{Acknowledgments}

This work has been partially supported by a grant to JRG from the Natural Sciences and Engineering Research Council of Canada.

\ifCLASSOPTIONcaptionsoff
  \newpage
\fi


\bibliographystyle{IEEEtran}
\bibliography{references.bib}

\begin{thebibliography}{10}
\providecommand{\url}[1]{#1}
\csname url@samestyle\endcsname
\providecommand{\newblock}{\relax}
\providecommand{\bibinfo}[2]{#2}
\providecommand{\BIBentrySTDinterwordspacing}{\spaceskip=0pt\relax}
\providecommand{\BIBentryALTinterwordstretchfactor}{4}
\providecommand{\BIBentryALTinterwordspacing}{\spaceskip=\fontdimen2\font plus
\BIBentryALTinterwordstretchfactor\fontdimen3\font minus
  \fontdimen4\font\relax}
\providecommand{\BIBforeignlanguage}[2]{{%
\expandafter\ifx\csname l@#1\endcsname\relax
\typeout{** WARNING: IEEEtran.bst: No hyphenation pattern has been}%
\typeout{** loaded for the language `#1'. Using the pattern for}%
\typeout{** the default language instead.}%
\else
\language=\csname l@#1\endcsname
\fi
#2}}
\providecommand{\BIBdecl}{\relax}
\BIBdecl

\bibitem{lewis2011network}
T.~G. Lewis, \emph{Network science: Theory and applications}.\hskip 1em plus
  0.5em minus 0.4em\relax John Wiley \& Sons, 2011.

\bibitem{van2010graph}
M.~Van~Steen, ``Graph theory and complex networks,'' \emph{An introduction},
  vol. 144, 2010.

\bibitem{cordeiro2018evolving}
M.~Cordeiro, R.~P. Sarmento, P.~Brazdil, and J.~Gama, ``Evolving networks and
  social network analysis methods and techniques,'' \emph{Social media and
  journalism-trends, connections, implications}, pp. 101--134, 2018.

\bibitem{handcock2003statistical}
M.~S. Handcock, \emph{Statistical models for social networks: Inference and
  degeneracy}.\hskip 1em plus 0.5em minus 0.4em\relax na, 2003.

\bibitem{albert2002statistical}
R.~Albert and A.-L. Barab{\'a}si, ``Statistical mechanics of complex
  networks,'' \emph{Reviews of modern physics}, vol.~74, no.~1, p.~47, 2002.

\bibitem{lu2011link}
L.~L{\"u} and T.~Zhou, ``Link prediction in complex networks: A survey,''
  \emph{Physica A: statistical mechanics and its applications}, vol. 390,
  no.~6, pp. 1150--1170, 2011.

\bibitem{wang2005group}
X.~Wang, N.~Mohanty, and A.~McCallum, ``Group and topic discovery from
  relations and text,'' in \emph{Proceedings of the 3rd international workshop
  on Link discovery}, 2005, pp. 28--35.

\bibitem{getoor2007markov}
L.~Getoor and B.~Taskar, ``Markov logic: A unifying framework for statistical
  relational learning,'' 2007.

\bibitem{getoor2005link}
L.~Getoor and C.~P. Diehl, ``Link mining: a survey,'' \emph{Acm Sigkdd
  Explorations Newsletter}, vol.~7, no.~2, pp. 3--12, 2005.

\bibitem{bell2007lessons}
R.~M. Bell and Y.~Koren, ``Lessons from the netflix prize challenge,''
  \emph{Acm Sigkdd Explorations Newsletter}, vol.~9, no.~2, pp. 75--79, 2007.

\bibitem{kazemi2018simple}
S.~M. Kazemi and D.~Poole, ``Simple embedding for link prediction in knowledge
  graphs,'' \emph{Advances in neural information processing systems}, vol.~31,
  2018.

\bibitem{perozzi2014deepwalk}
B.~Perozzi, R.~Al-Rfou, and S.~Skiena, ``Deepwalk: Online learning of social
  representations,'' in \emph{Proceedings of the 20th ACM SIGKDD international
  conference on Knowledge discovery and data mining}, 2014, pp. 701--710.

\bibitem{grover2016node2vec}
A.~Grover and J.~Leskovec, ``node2vec: Scalable feature learning for
  networks,'' in \emph{Proceedings of the 22nd ACM SIGKDD international
  conference on Knowledge discovery and data mining}, 2016, pp. 855--864.

\bibitem{zhang2018end}
M.~Zhang, Z.~Cui, M.~Neumann, and Y.~Chen, ``An end-to-end deep learning
  architecture for graph classification,'' in \emph{Thirty-second AAAI
  conference on artificial intelligence}, 2018.

\bibitem{zhang2018link}
M.~Zhang and Y.~Chen, ``Link prediction based on graph neural networks,''
  \emph{Advances in neural information processing systems}, vol.~31, 2018.

\bibitem{berlusconi2016link}
G.~Berlusconi, F.~Calderoni, N.~Parolini, M.~Verani, and C.~Piccardi, ``Link
  prediction in criminal networks: A tool for criminal intelligence analysis,''
  \emph{PloS one}, vol.~11, no.~4, p. e0154244, 2016.

\bibitem{wolpert1992stacked}
D.~H. Wolpert, ``Stacked generalization,'' \emph{Neural networks}, vol.~5,
  no.~2, pp. 241--259, 1992.

\bibitem{ghasemian2020stacking}
A.~Ghasemian, H.~Hosseinmardi, A.~Galstyan, E.~M. Airoldi, and A.~Clauset,
  ``Stacking models for nearly optimal link prediction in complex networks,''
  \emph{Proceedings of the National Academy of Sciences}, vol. 117, no.~38, pp.
  23\,393--23\,400, 2020.

\bibitem{platt1999probabilistic}
J.~Platt \emph{et~al.}, ``Probabilistic outputs for support vector machines and
  comparisons to regularized likelihood methods,'' \emph{Advances in large
  margin classifiers}, vol.~10, no.~3, pp. 61--74, 1999.

\bibitem{zadrozny2002transforming}
B.~Zadrozny and C.~Elkan, ``Transforming classifier scores into accurate
  multiclass probability estimates,'' in \emph{Proceedings of the eighth ACM
  SIGKDD international conference on Knowledge discovery and data mining},
  2002, pp. 694--699.

\bibitem{tabacof2019probability}
P.~Tabacof and L.~Costabello, ``Probability calibration for knowledge graph
  embedding models,'' in \emph{International Conference on Learning
  Representations}, 2019.

\bibitem{wang2021confident}
X.~Wang, H.~Liu, C.~Shi, and C.~Yang, ``Be confident! towards trustworthy graph
  neural networks via confidence calibration,'' \emph{Advances in Neural
  Information Processing Systems}, vol.~34, 2021.

\bibitem{gosak2018network}
M.~Gosak, R.~Markovi{\v{c}}, J.~Dolen{\v{s}}ek, M.~S. Rupnik, M.~Marhl,
  A.~Sto{\v{z}}er, and M.~Perc, ``Network science of biological systems at
  different scales: A review,'' \emph{Physics of life reviews}, vol.~24, pp.
  118--135, 2018.

\bibitem{dick2018reciprocal}
K.~Dick and J.~R. Green, ``Reciprocal perspective for improved protein-protein
  interaction prediction,'' \emph{Scientific reports}, vol.~8, no.~1, pp.
  1--12, 2018.

\bibitem{dick2020pipe4}
K.~Dick, B.~Samanfar, B.~Barnes, E.~R. Cober, B.~Mimee, L.~H. Tan, S.~J.
  Molnar, K.~K. Biggar, A.~Golshani, F.~Dehne \emph{et~al.}, ``Pipe4: Fast ppi
  predictor for comprehensive inter-and cross-species interactomes,''
  \emph{Scientific reports}, vol.~10, no.~1, pp. 1--15, 2020.

\bibitem{kyrollos2020rpmirdip}
D.~G. Kyrollos, B.~Reid, K.~Dick, and J.~R. Green, ``Rpmirdip: Reciprocal
  perspective improves mirna targeting prediction,'' \emph{Scientific reports},
  vol.~10, no.~1, pp. 1--13, 2020.

\bibitem{musdti}
K.~Dick, D.~G. Kyrollos, E.~D. Cosoreanu, J.~Dooley, J.~S. Fryer, S.~M. Gordon,
  N.~Kharbanda, M.~Klamrowski, P.~N.~L. LaCasse, T.~F. Leung, M.~A. Nasir,
  C.~Qiu, A.~S. Robinson, D.~Shao, B.~R. Siromahov, E.~Starlight, C.~Tran,
  C.~Wang, Y.-K. Yang, and J.~R. Green, ``Musdti: Meta undergraduate student
  drug-target interaction predictor,'' \emph{Scientific reports}, 2022. Under
  Review.

\bibitem{dick2021multi}
K.~Dick, A.~Chopra, K.~K. Biggar, and J.~R. Green, ``Multi-schema computational
  prediction of the comprehensive sars-cov-2 vs. human interactome,''
  \emph{PeerJ}, vol.~9, p. e11117, 2021.

\bibitem{kashima2009link}
H.~Kashima, T.~Kato, Y.~Yamanishi, M.~Sugiyama, and K.~Tsuda, ``Link
  propagation: A fast semi-supervised learning algorithm for link prediction,''
  in \emph{Proceedings of the 2009 SIAM international conference on data
  mining}.\hskip 1em plus 0.5em minus 0.4em\relax SIAM, 2009, pp. 1100--1111.

\bibitem{shahreza2017heter}
M.~L. Shahreza, N.~Ghadiri, S.~R. Mousavi, J.~Varshosaz, and J.~R. Green,
  ``Heter-lp: A heterogeneous label propagation algorithm and its application
  in drug repositioning,'' \emph{Journal of biomedical informatics}, vol.~68,
  pp. 167--183, 2017.

\bibitem{berton2015link}
L.~Berton, J.~Valverde-Rebaza, and A.~de~Andrade~Lopes, ``Link prediction in
  graph construction for supervised and semi-supervised learning,'' in
  \emph{2015 International Joint Conference on Neural Networks (IJCNN)}.\hskip
  1em plus 0.5em minus 0.4em\relax IEEE, 2015, pp. 1--8.

\bibitem{ceci2015semi}
M.~Ceci, G.~Pio, V.~Kuzmanovski, and S.~D{\v{z}}eroski, ``Semi-supervised
  multi-view learning for gene network reconstruction,'' \emph{PloS one},
  vol.~10, no.~12, p. e0144031, 2015.

\bibitem{Shi2018ECCV}
W.~Shi, Y.~Gong, C.~Ding, Z.~M. Tao, and N.~Zheng, ``Transductive
  semi-supervised deep learning using min-max features,'' in \emph{Proceedings
  of the European Conference on Computer Vision (ECCV)}, September 2018.

\bibitem{Song2018CVPR}
J.~Song, C.~Shen, Y.~Yang, Y.~Liu, and M.~Song, ``Transductive unbiased
  embedding for zero-shot learning,'' in \emph{Proceedings of the IEEE
  Conference on Computer Vision and Pattern Recognition (CVPR)}, June 2018.

\bibitem{yones2018genome}
C.~Yones, G.~Stegmayer, and D.~H. Milone, ``Genome-wide pre-mirna discovery
  from few labeled examples,'' \emph{Bioinformatics}, vol.~34, no.~4, pp.
  541--549, 2018.

\bibitem{daud2017will}
A.~Daud, W.~Ahmed, T.~Amjad, J.~A. Nasir, N.~R. Aljohani, R.~A. Abbasi, and
  I.~Ahmad, ``Who will cite you back? reciprocal link prediction in citation
  networks,'' \emph{Library Hi Tech}, 2017.

\bibitem{li2020link}
J.~Li, J.~Peng, S.~Liu, X.~Ji, X.~Li, and X.~Hu, ``Link prediction in directed
  networks utilizing the role of reciprocal links,'' \emph{IEEE Access},
  vol.~8, pp. 28\,668--28\,680, 2020.

\bibitem{koren2009matrix}
Y.~Koren, R.~Bell, and C.~Volinsky, ``Matrix factorization techniques for
  recommender systems,'' \emph{Computer}, vol.~42, no.~8, pp. 30--37, 2009.

\bibitem{niculescu2005predicting}
A.~Niculescu-Mizil and R.~Caruana, ``Predicting good probabilities with
  supervised learning,'' in \emph{Proceedings of the 22nd international
  conference on Machine learning}, 2005, pp. 625--632.

\bibitem{harper2015movielens}
F.~M. Harper and J.~A. Konstan, ``The movielens datasets: History and
  context,'' \emph{Acm transactions on interactive intelligent systems (tiis)},
  vol.~5, no.~4, pp. 1--19, 2015.

\bibitem{vinh2020hyperml}
L.~Vinh~Tran, Y.~Tay, S.~Zhang, G.~Cong, and X.~Li, ``Hyperml: A boosting
  metric learning approach in hyperbolic space for recommender systems,'' in
  \emph{Proceedings of the 13th International Conference on Web Search and Data
  Mining}, 2020, pp. 609--617.

\bibitem{sun2022neural}
X.~Sun, L.~Gong, Z.~Han, P.~Zhao, J.~Yu, and S.~Wang, ``Neural metric
  factorization for recommendation,'' \emph{Mathematics}, vol.~10, no.~3, p.
  503, 2022.

\bibitem{slokom2018comparing}
M.~Slokom, ``Comparing recommender systems using synthetic data,'' in
  \emph{Proceedings of the 12th ACM Conference on Recommender Systems}, 2018,
  pp. 548--552.

\bibitem{he2016ups}
R.~He and J.~McAuley, ``Ups and downs: Modeling the visual evolution of fashion
  trends with one-class collaborative filtering,'' in \emph{proceedings of the
  25th international conference on world wide web}, 2016, pp. 507--517.

\bibitem{mcauley2015image}
J.~McAuley, C.~Targett, Q.~Shi, and A.~Van Den~Hengel, ``Image-based
  recommendations on styles and substitutes,'' in \emph{Proceedings of the 38th
  international ACM SIGIR conference on research and development in information
  retrieval}, 2015, pp. 43--52.

\bibitem{chen2015xgboost}
T.~Chen, T.~He, M.~Benesty, V.~Khotilovich, Y.~Tang, H.~Cho, K.~Chen
  \emph{et~al.}, ``Xgboost: extreme gradient boosting,'' \emph{R package
  version 0.4-2}, vol.~1, no.~4, pp. 1--4, 2015.

\bibitem{chapelle2006semi}
O.~Chapelle, B.~Scholkopf, and A.~Zien, ``Semi-supervised learning,''
  \emph{IEEE Transactions on Neural Networks}, vol.~20, no.~3, pp. 542--542,
  2006.

\bibitem{laine2016temporal}
S.~Laine and T.~Aila, ``Temporal ensembling for semi-supervised learning,''
  \emph{arXiv preprint arXiv:1610.02242}, 2016.

\bibitem{tarvainen2017mean}
A.~Tarvainen and H.~Valpola, ``Mean teachers are better role models:
  Weight-averaged consistency targets improve semi-supervised deep learning
  results,'' in \emph{Advances in neural information processing systems}, 2017,
  pp. 1195--1204.

\bibitem{krizhevsky2014cifar}
A.~Krizhevsky, V.~Nair, and G.~Hinton, ``The cifar-10 dataset,'' \emph{online:
  http://www. cs. toronto. edu/kriz/cifar. html}, 2014.

\bibitem{sksurprise}
\BIBentryALTinterwordspacing
N.~Hug, ``Surprise: A python library for recommender systems,'' \emph{Journal
  of Open Source Software}, vol.~5, no.~52, p. 2174, 2020. [Online]. Available:
  \url{https://doi.org/10.21105/joss.02174}
\BIBentrySTDinterwordspacing

\bibitem{aggarwal2016recommender}
C.~C. Aggarwal \emph{et~al.}, \emph{Recommender systems}.\hskip 1em plus 0.5em
  minus 0.4em\relax Springer, 2016, vol.~1.

\bibitem{luo2014efficient}
X.~Luo, M.~Zhou, Y.~Xia, and Q.~Zhu, ``An efficient non-negative
  matrix-factorization-based approach to collaborative filtering for
  recommender systems,'' \emph{IEEE Transactions on Industrial Informatics},
  vol.~10, no.~2, pp. 1273--1284, 2014.

\bibitem{lemire2005slope}
D.~Lemire and A.~Maclachlan, ``Slope one predictors for online rating-based
  collaborative filtering,'' in \emph{Proceedings of the 2005 SIAM
  International Conference on Data Mining}.\hskip 1em plus 0.5em minus
  0.4em\relax SIAM, 2005, pp. 471--475.

\bibitem{george2005scalable}
T.~George and S.~Merugu, ``A scalable collaborative filtering framework based
  on co-clustering,'' in \emph{Fifth IEEE International Conference on Data
  Mining (ICDM'05)}.\hskip 1em plus 0.5em minus 0.4em\relax IEEE, 2005, pp.
  4--pp.

\bibitem{wilcoxon1992individual}
F.~Wilcoxon, ``Individual comparisons by ranking methods,'' in
  \emph{Breakthroughs in statistics}.\hskip 1em plus 0.5em minus 0.4em\relax
  Springer, 1992, pp. 196--202.

\bibitem{feuerverger2012statistical}
A.~Feuerverger, Y.~He, and S.~Khatri, ``Statistical significance of the netflix
  challenge,'' \emph{Statistical Science}, vol.~27, no.~2, pp. 202--231, 2012.

\bibitem{dick2018fitting}
K.~Dick and J.~R. Green, ``Fitting rank order data in the age of context,'' in
  \emph{2018 IEEE Life Sciences Conference (LSC)}.\hskip 1em plus 0.5em minus
  0.4em\relax IEEE, 2018, pp. 142--146.

\end{thebibliography}

\vspace{5mm} 

\begin{IEEEbiography}[{\includegraphics[width=1in,height=1.25in,clip,keepaspectratio]{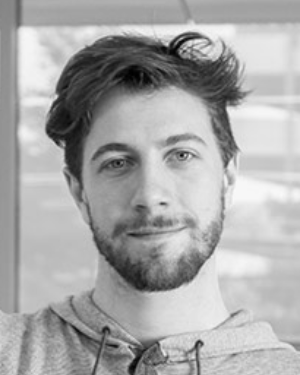}}]{Kevin Dick} (M'15) received the B.Sc. degree in biology and computer science from McGill University, Montréal, QC, Canada in 2014 and is currently pursuing a Ph.D. degree in biomedical engineering (specialized in data science and bioinformatics) at Carleton University, Ottawa, ON, Canada.

He is currently a Research Assistant at the Carleton University Biomedical Informatics Collaboratory (cuBIC) and a Research Scientist for the Centre of Access to Information and Justice (CAIJ) at the University of Winnipeg. His current research includes machine learning and the use of high-performance computing infrastructure for biomedical informatic applications, computer vision to study phenomena of the natural environment, and the application of computational methods to solve digital access challenges.
\end{IEEEbiography}

\begin{IEEEbiography}[{\includegraphics[width=1in,height=1.25in,clip,keepaspectratio]{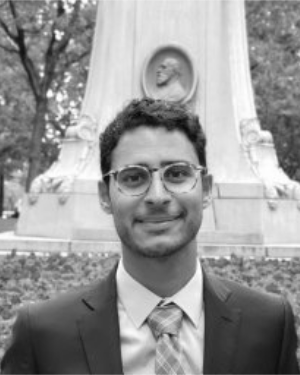}}]{Daniel G. Kyrollos}  (S'2020) received his B.Eng degree in Biomedical and Electrical Engineering from Carleton. He is currently pursuing a Master of Applied Science in Electrical and Computer Engineering with a specialization in Data Science at Carleton University. His research interests include data science, machine learning, machine vision, natural language processing and signal processing. Currently his thesis project aims to investigate novel patient monitoring technologies in the neonatal intensive care unit. The goal is to examine the use of pressure-sensitive mats, video data and depth data to provide continuous, unobtrusive, and non-contact monitoring of critically ill babies.

\end{IEEEbiography}

\begin{IEEEbiography}[{\includegraphics[width=1in,height=1.25in,clip,keepaspectratio]{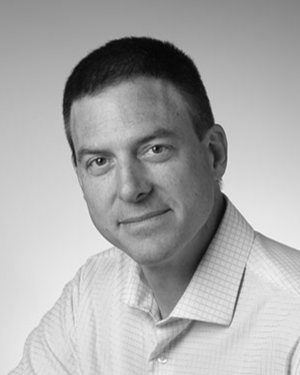}}]{James R. Green} (S’02–M’05–SM’10) received the B.A.Sc. degree in systems design engineering from the University of Waterloo, Waterloo, ON, Canada, in 1998, and the M.A.Sc. and Ph.D. degrees in electrical and computer engineering from Queen’s University, Kingston, ON, Canada, in 2000 and 2005, respectively.

He is currently a Professor with the Department of Systems and Computer Engineering, Carleton University, Ottawa, ON, Canada. His research interests include a range of machine learning challenges within biomedical informatics, with focus on problems exhibiting class imbalance.
\end{IEEEbiography}




\end{document}


\title{Supplementary Materials\\\Large Generalized Reciprocal Perspective}

\author{Kevin~Dick,~\IEEEmembership{Graduate~Student~Member,~IEEE,}
 Daniel~G.~Kyrollos,~\IEEEmembership{Graduate~Student~Member,~IEEE}
 and~James~R.~Green,~\IEEEmembership{Senior~Member,~IEEE}
\IEEEcompsocitemizethanks{\IEEEcompsocthanksitem K. Dick, D. G. Kyrollos, and J. R. Green are with the Department of Systems \& Computer Engineering, Carleton University, Ottawa, Ontario, Canada.\protect\\
E-mail: kevin.dick@carleton.ca

\IEEEcompsocthanksitem The authors are also members of the Institute of Data Science, Carleton University, Ottawa, Canada.}
\thanks{}}

\markboth{}%
{Dick \MakeLowercase{\textit{et al.}}: Supplementary Materials - Reciprocal Perspective}

\IEEEtitleabstractindextext{%
\begin{abstract}

These supplementary materials in accompaniment to the generalized Reciprocal Perspective manuscript contains materials important to the experimental design. Specifically, these materials illustrate many of the visualized intermediary and final findings relevant to the Amazon Ratings Dataset manipulation. 


\end{abstract}

\begin{IEEEkeywords}
Recommendation Systems, Semi-Supervised Learning, Class Imbalance
\end{IEEEkeywords}}

\maketitle
\IEEEdisplaynontitleabstractindextext
\IEEEpeerreviewmaketitle

\IEEEraisesectionheading{\section{Introduction}\label{sec:introduction}}
\beginsupplement
\IEEEPARstart{T}{he} following supplementary materials contain various visualizations and tabulations of in support of the generalized Reciprocal Perspective (RP) manuscript. This work draws from the numerous studies on RP in the bioinformatic domain \cite{dick2018reciprocal, dick2018fitting, dick2020pipe4, kyrollos2020rpmirdip, dick2021multi}. Numerous experiments were performed to manipulate the Amazon Ratings dataset and the tables and figures below illustrate these Amazon resizing experiments. 

In Table \ref{tab:amazon-sizes} we summarize the expected computational requirements to naively applying RP to each product sub-category of this dataset. Most concerning is the magnitude of each of the CPM sizes when stored in RAM or on disk. Even the smallest CPM (Musical Instruments) would require approximately 678 GB of RAM to be stored in memory. On the largest end, the Books category would necessitate a CPM of 722.6 TB in size. Clearly the compute infrastructure and algorithmic consideration required to generate and manipulate these matricies is possible only in the most extreme cases.

\begin{figure*}[p]
 \centering
 \includegraphics[width=0.95\textwidth]{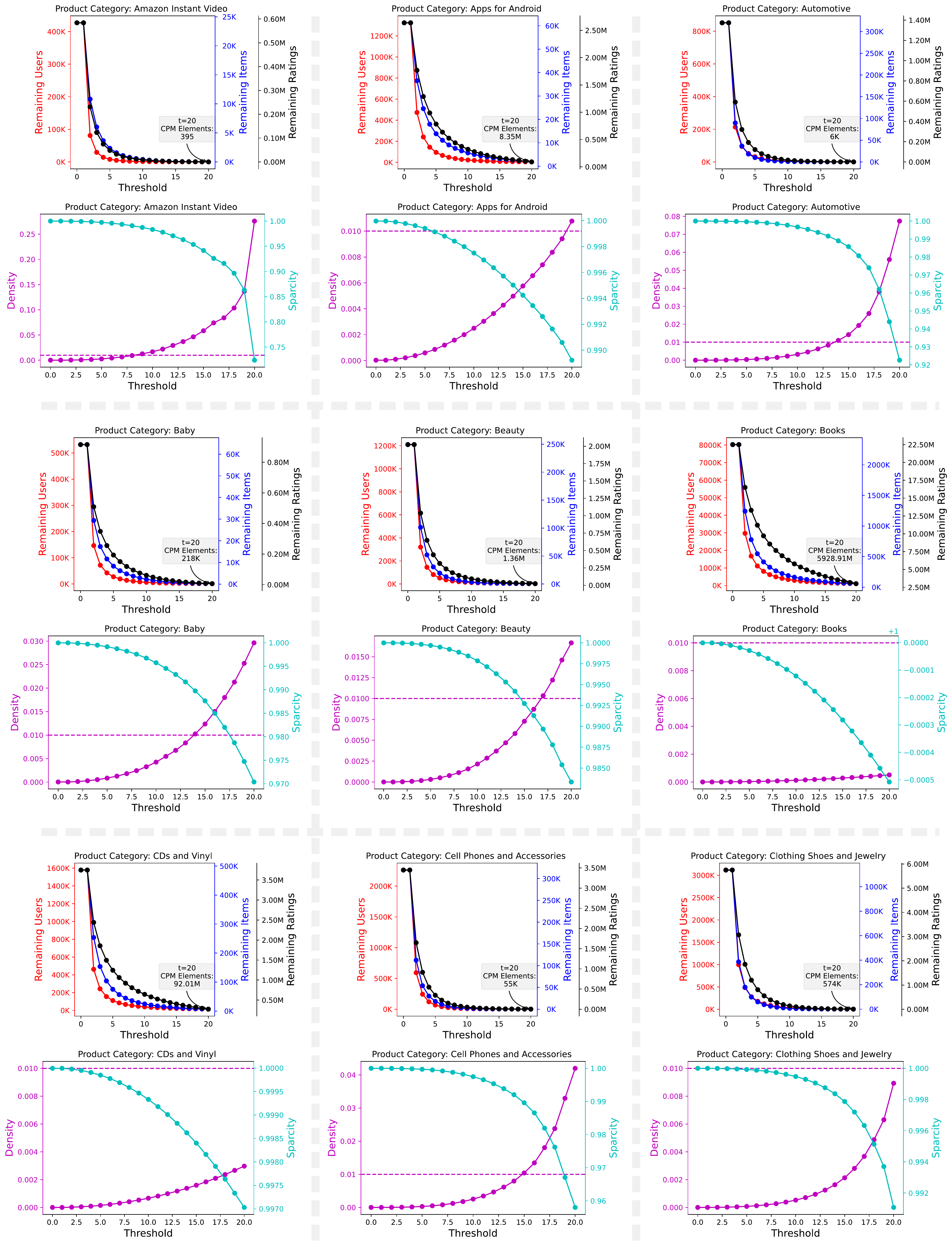}
 \caption{Amazon Product Category Matrix Resizing by Thresholding Long-Tail Rating Distributions.}
 \label{fig:thresh1}
\end{figure*}

\begin{figure*}[p]
 \centering
 \includegraphics[width=0.95\textwidth]{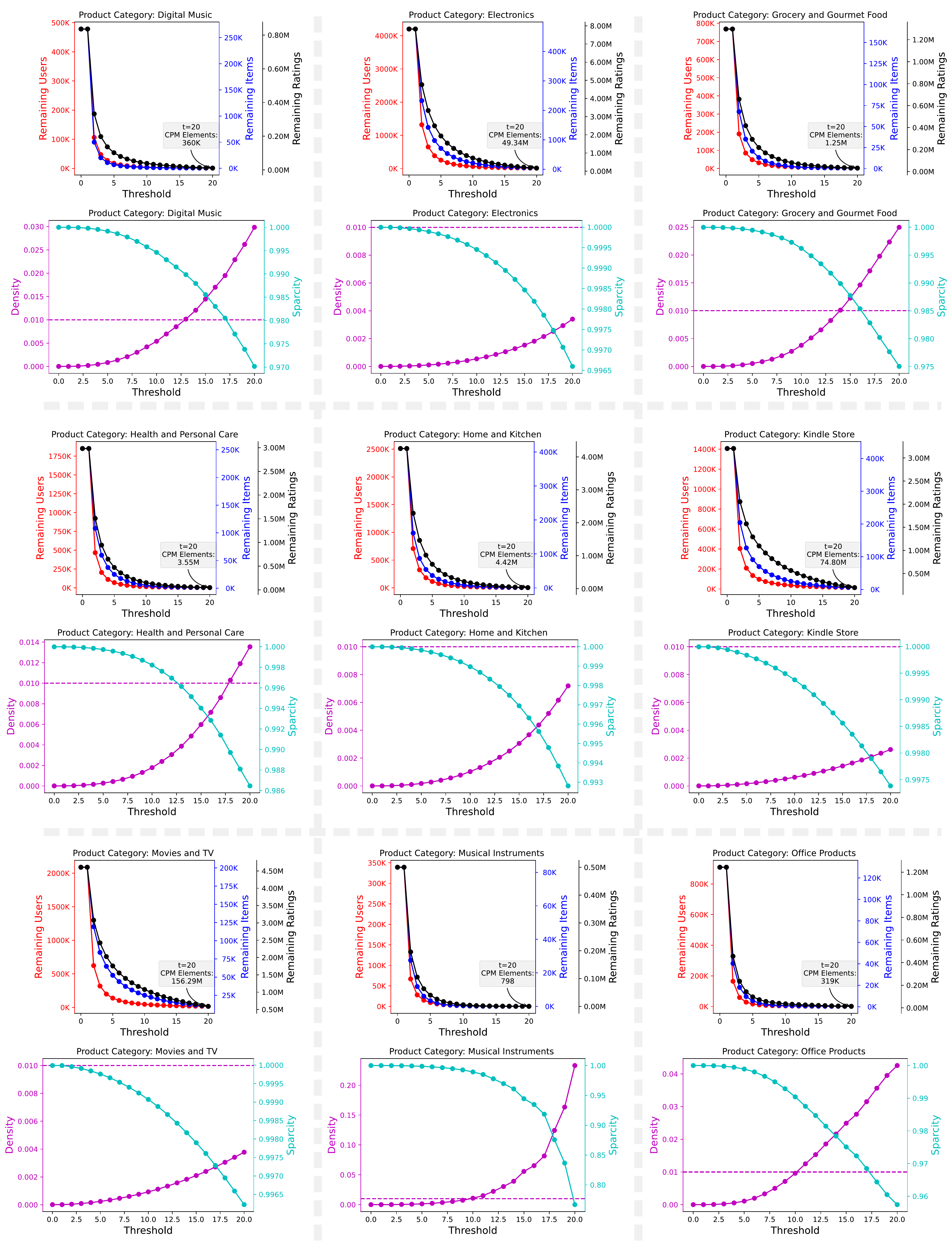}
 \caption{Amazon Product Category Matrix Resizing by Thresholding Long-Tail Rating Distributions.}
 \label{fig:thresh2}
\end{figure*}

\begin{figure*}[p]
 \centering
 \includegraphics[width=0.95\textwidth]{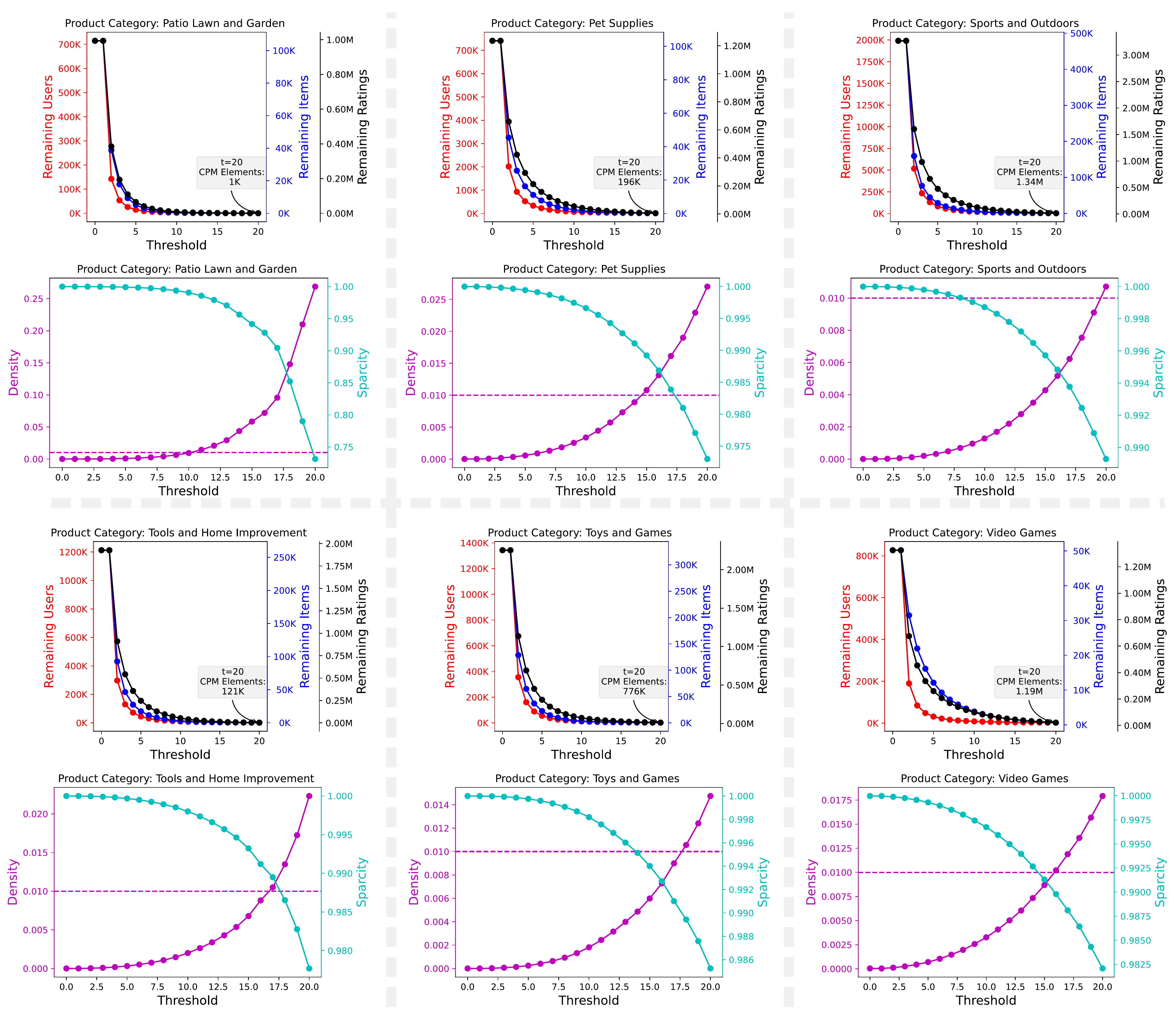}
 \caption{Amazon Product Category Matrix Resizing by Thresholding Long-Tail Rating Distributions.}
 \label{fig:thresh3}
\end{figure*}

Given that the purpose of this work is to demonstrate the broad applicability of RP to \textit{reasonably sized} problems, we sought to reduce the CPM size using a few strategies. Given the long-tailed distributions of ratings data (the vaste majority of users rate only a few items and the vaste majority of items are rated by only a few users), the typical approach is to extract the $k$-core of ratings. That is, only the users and items that provided and recieved at least $k$ ratings are considered. Thus, we explored the trade-off in applying a constant threshold $k$ to extract the $k$-core from each of the 24 Amazon product categories (Figs. \ref{fig:thresh1}, \ref{fig:thresh2}, \ref{fig:thresh3}). 

While a $k=20$ appeared reasonable for the majority of the product sub-categories, the smallest and largest CPMs seemed either unreasonably small or still overwhelmingly large. Consequently, we chose to vary the threshold for the $k$-core on a per-dataset basis by keeping an alternative metric constant. We chose to use a density index of 1\% given that this (yet) represented a far more sparse input dataset than the others considered in this study while also enabling the cross-category generation of the CPMs for subsequent RP application.

Ultimately, the Amazon Product rating datasets restricted to spaces constrained by 1\% density index were leveraged in this work and consistent RP results were reported. We thought that in varying the constrained density index by another order of magnitude, 10\%, might yield interesting results, however, as summarized in Table \ref{tab:amz-10PDensities}, the application of this constraint is clearly too severe across all Amazon Product categories given that too few users and item remain. 






\begin{landscape}
\centering
\begin{table}[p]
\caption{Amazon Product Category Rating Data and Anticipated Compute Requirements for CPM Generation.}
\label{tab:amazon-sizes}
\resizebox{1.3\textwidth}{!}{
\begin{tabular}{lccccccc}
\hline
\begin{tabular}[c]{@{}l@{}}Amazon Product\\ Category\end{tabular} & \begin{tabular}[c]{@{}c@{}}Num. \\ Ratings\end{tabular} & \multicolumn{1}{c}{\begin{tabular}[c]{@{}c@{}}Num. \\ Users\end{tabular}} & \begin{tabular}[c]{@{}c@{}}Num. \\ Items\end{tabular} & \begin{tabular}[c]{@{}c@{}}CPM Matrix \\ Elements\end{tabular} & \begin{tabular}[c]{@{}c@{}}CPM Size \\ in Req. \\ RAM (TBs)\end{tabular} & \begin{tabular}[c]{@{}c@{}}Density\\ Index\end{tabular} & \begin{tabular}[c]{@{}c@{}}Sparcity\\ Index\end{tabular} \\ \hline
\multicolumn{1}{l|}{Musical Instruments} & 500,176 & 339,231 & 83,046  & 169,675,204,656  & 0.6787  & 1.78E-05  & 0.999982245 \\
\multicolumn{1}{l|}{Amazon Instant Video} & 583,933 & 426,922 & 23,965  & 249,293,844,226  & 0.9972  & 5.71E-05  & 0.999942926 \\
\multicolumn{1}{l|}{Digital Music}  & 836,006 & 478,235 & 266,414  & 399,807,329,410  & 1.5992  & 6.56E-06  & 0.999993438 \\
\multicolumn{1}{l|}{Baby} & 915,446 & 531,890 & 64,426  & 486,916,572,940  & 1.9477  & 2.67E-05  & 0.999973285 \\
\multicolumn{1}{l|}{Patio Lawn and Garden}  & 993,490 & 714,791 & 105,984  & 710,137,710,590  & 2.8406  & 1.31E-05  & 0.999986886 \\
\multicolumn{1}{l|}{Pet Supplies}   & 1,235,316  & 740,985 & 103,288  & 915,350,626,260  & 3.6614  & 1.61E-05  & 0.999983859 \\
\multicolumn{1}{l|}{Grocery and Gourmet\_Food} & 1,297,156  & 768,438 & 166,049  & 996,783,962,328  & 3.9871  & 1.02E-05  & 0.999989834 \\
\multicolumn{1}{l|}{Video Games} & 1,324,753  & 826,767 & 50,210  & 1,095,262,063,551 & 4.3810  & 3.19E-05  & 0.999968087 \\
\multicolumn{1}{l|}{Office Products} & 1,243,186  & 909,314 & 130,006  & 1,130,446,434,404 & 4.5218  & 1.05E-05  & 0.999989484 \\
\multicolumn{1}{l|}{Automotive}  & 1,373,768  & 851,418 & 320,112  & 1,169,650,803,024 & 4.6786  & 5.04E-06  & 0.99999496  \\
\multicolumn{1}{l|}{Tools and Home Improvement}  & 1,926,047  & 1,212,468  & 260,659  & 2,335,270,353,996 & 9.3411  & 6.09E-06  & 0.999993906 \\
\multicolumn{1}{l|}{Beauty}  & 2,023,070  & 1,210,271  & 249,274  & 2,448,462,951,970 & 9.7939  & 6.71E-06  & 0.999993294 \\
\multicolumn{1}{l|}{Toys and Games}  & 2,252,771  & 1,342,911  & 327,698  & 3,025,270,956,381 & 12.1011  & 5.12E-06  & 0.999994881 \\
\multicolumn{1}{l|}{Apps for Android} & 2,638,172  & 1,323,884  & 61,275  & 3,492,633,700,048 & 13.9705  & 3.25E-05  & 0.999967479 \\
\multicolumn{1}{l|}{Kindle Store}   & 3,205,467  & 1,406,890  & 430,530  & 4,509,739,467,630 & 18.0390  & 5.29E-06  & 0.999994708 \\
\multicolumn{1}{l|}{Health and Personal Care} & 2,982,326  & 1,851,132  & 252,331  & 5,520,679,093,032 & 22.0827  & 6.38E-06  & 0.999993615 \\
\multicolumn{1}{l|}{CDs and Vinyl}  & 3,749,004  & 1,578,597  & 486,360  & 5,918,166,467,388 & 23.6727  & 4.88E-06  & 0.999995117 \\
\multicolumn{1}{l|}{Sports and Outdoors} & 3,268,695  & 1,990,521  & 478,898  & 6,506,406,040,095 & 26.0256  & 3.43E-06  & 0.999996571 \\
\multicolumn{1}{l|}{Cell Phones and Accessories}  & 3,447,249  & 2,261,045  & 319,678  & 7,794,385,115,205 & 31.1775  & 4.77E-06  & 0.999995231 \\
\multicolumn{1}{l|}{Movies and TV}  & 4,607,047  & 2,088,620  & 200,941  & 9,622,370,505,140 & 38.4895  & 1.10E-05  & 0.999989023 \\
\multicolumn{1}{l|}{Home and Kitchen} & 4,253,926  & 2,511,610  & 410,243  & 10,684,203,080,860 & 42.7368  & 4.13E-06  & 0.999995871 \\
\multicolumn{1}{l|}{Clothing Shoes and Jewelry}  & 5,748,920  & 3,117,268  & 1,136,004 & 17,920,924,350,560 & 71.6837  & 1.62E-06  & 0.999998377 \\
\multicolumn{1}{l|}{Electronics} & 7,824,482  & 4,201,696  & 476,002  & 32,876,094,721,472 & 131.5044 & 3.91E-06  & 0.999996088 \\
\multicolumn{1}{l|}{Books} & 22,507,155 & 8,026,324  & 2,330,066 & 180,649,718,348,220 & 722.5989 & 1.20E-06  & 0.999998797 \\ \hline
\end{tabular}}
\end{table}
\end{landscape}

\begin{table*}[]
  \centering
  \caption{Summary of Recommendation System RMSE Performance over Three Ratings Benchmark Datasets.}
  \includegraphics[width=\textwidth]{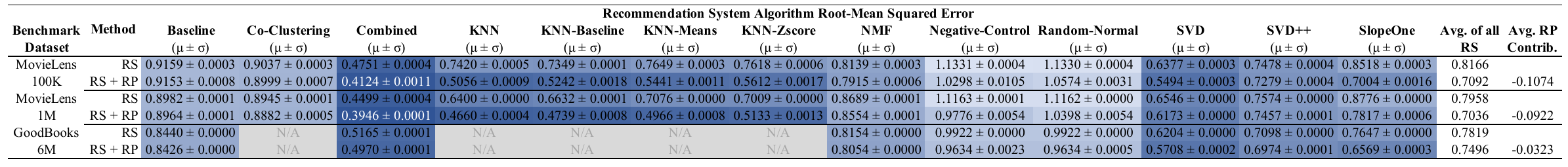}
  \label{tab:result-mlgb}
\end{table*}

\begin{table*}[p]
  \caption{Summary of Recommendation System RMSE Performance on each Amazon Product Sub-Category from Applying RP over the Consistent 1\% Matrix Densification Experiment. Cell colouring is applied table-wide to express global RMSE results, therefore, dataset-specific interpretation of results must be consider row-wise in pairs.}
  \label{tab:result-1PDensities}
  \includegraphics[width=\textwidth]{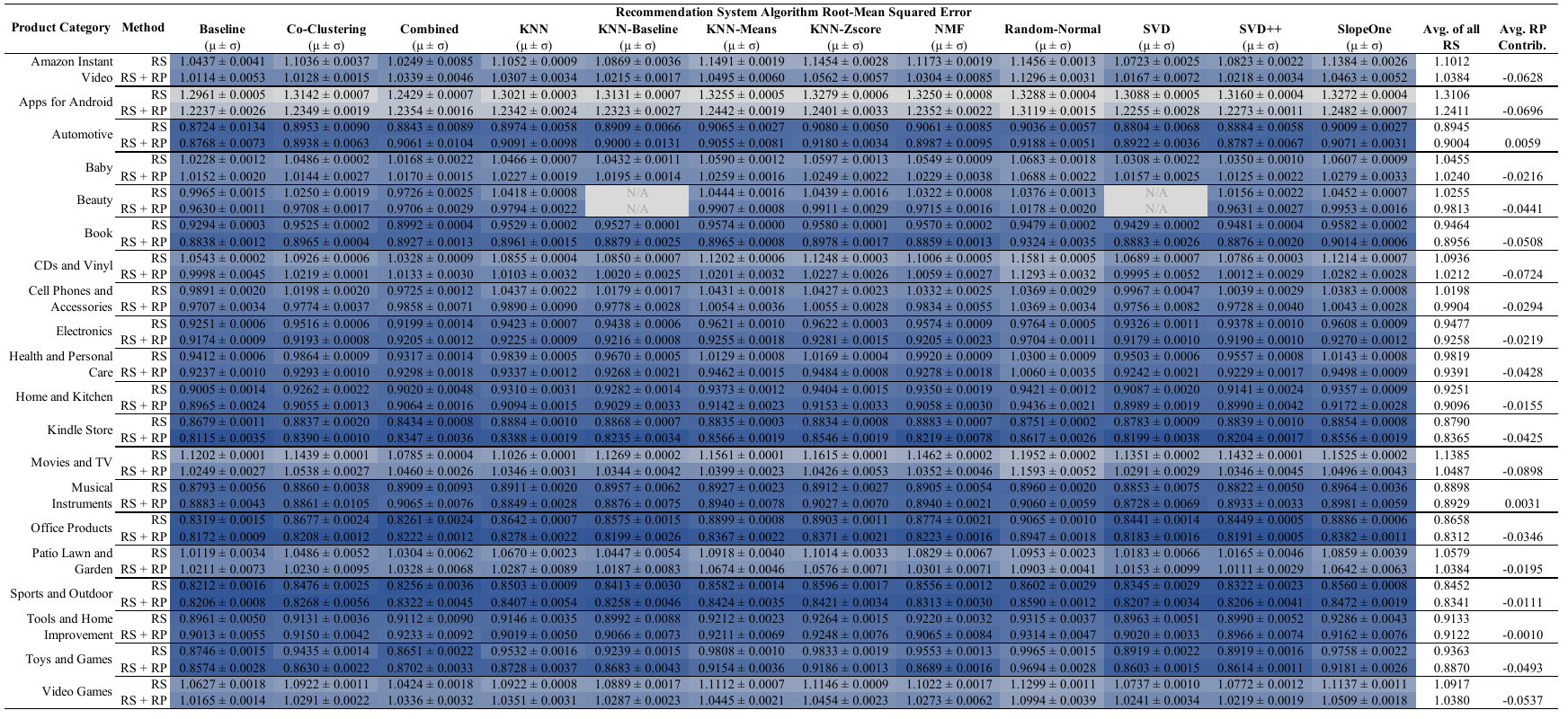}
\end{table*}

\begin{table*}[]
 \centering
 \caption{Summary of Recommendation System RMSE Performance on each Amazon Product Sub-Category from Applying RP over the Global t=20 Threshold Experiment.}
 \includegraphics[width=\textwidth]{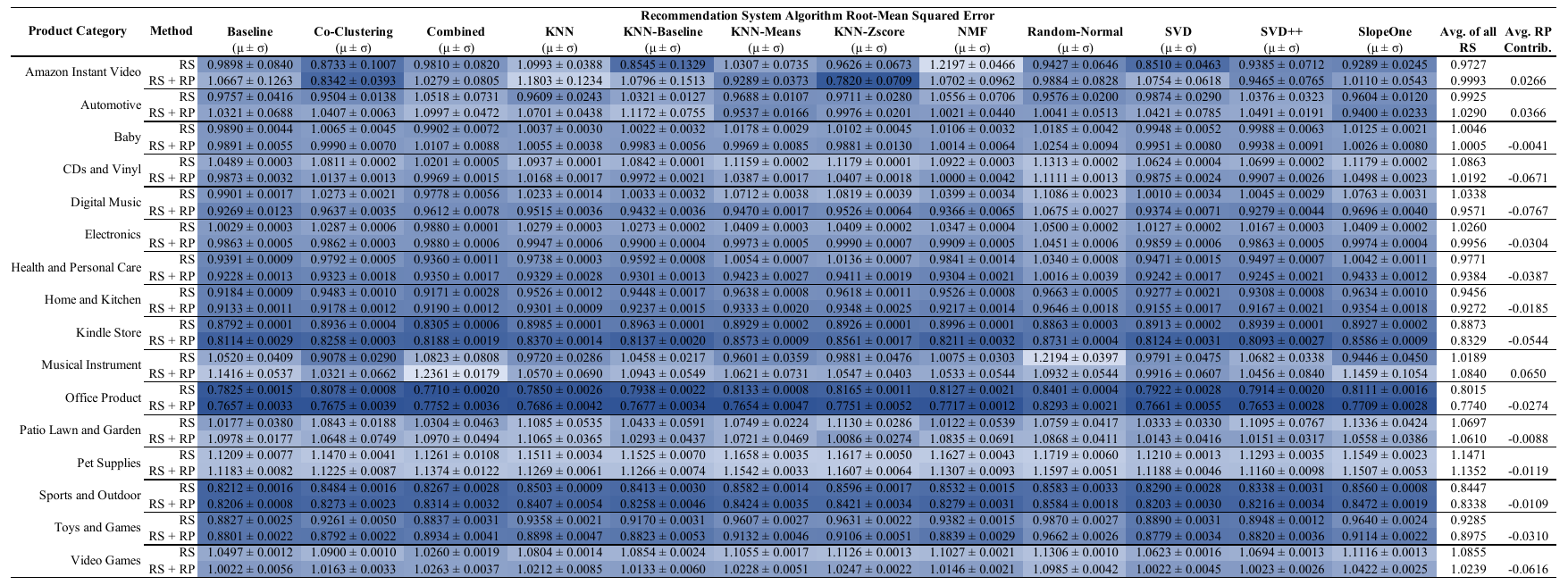}
 \label{tab:result-t20}
\end{table*}


\begin{table*}[]
\centering
\caption{Amazon Product Ratings Dataset Densification to 10\%}
\label{tab:amz-10PDensities}
\begin{tabular}{rccccc
>{\columncolor[HTML]{EFEFEF}}l l}
\toprule
\begin{tabular}[c]{@{}r@{}}Product\\ Category\end{tabular} & Threshold & \begin{tabular}[c]{@{}c@{}}Remaining\\ Users\end{tabular} & \begin{tabular}[c]{@{}c@{}}Remaining\\ Items\end{tabular} & \begin{tabular}[c]{@{}c@{}}Remaining\\ Ratings\end{tabular} & \begin{tabular}[c]{@{}c@{}}CPM\\ Elements\end{tabular} & Density & Sparcity \\ \midrule
\multicolumn{1}{r|}{\begin{tabular}[c]{@{}r@{}}Amazon Instant \\ Video\end{tabular}}  & 18  & 201  & 32 & 666 & 3,216 & 0.1035  & 0.8965 \\
\multicolumn{1}{r|}{Apps for Android}  & 44  & 622  & 49 & 3,096  & 15,239 & 0.1016  & 0.8984 \\
\multicolumn{1}{r|}{Automotive}  & 21  & 270  & 11 & 333 & 1,485 & 0.1121  & 0.8879 \\
\multicolumn{1}{r|}{Baby} & 28  & 322  & 19 & 677 & 3,059 & 0.1107  & 0.8893 \\
\multicolumn{1}{r|}{Beauty}  & 34  & 407  & 123  & 5,509  & 25,030 & 0.1100  & 0.8900 \\
\multicolumn{1}{r|}{Book} & 177 & 2,252 & 28 & 6,324  & 31,528 & 0.1003  & 0.8997 \\
\multicolumn{1}{r|}{CDs and Vinyl}  & 87  & 1,090 & 44 & 4,821  & 23,980 & 0.1005  & 0.8995 \\
\multicolumn{1}{r|}{\begin{tabular}[c]{@{}r@{}}Cell Phones and \\ Accessories\end{tabular}} & 24  & 257  & 27 & 859 & 3,469 & 0.1238  & 0.8762 \\
\multicolumn{1}{r|}{\begin{tabular}[c]{@{}r@{}}Clothing Shoes and \\ Jewelry\end{tabular}}  & 28  & 318  & 11 & 371 & 1,749 & 0.1061  & 0.8939 \\
\multicolumn{1}{r|}{Digital Music}  & 31  & 355  & 40 & 1,477  & 7,100 & 0.1040  & 0.8960 \\
\multicolumn{1}{r|}{Electronics} & 59  & 793  & 30 & 2,674  & 11,895 & 0.1124  & 0.8876 \\
\multicolumn{1}{r|}{\begin{tabular}[c]{@{}r@{}}Grocery and Gourmet \\ Food\end{tabular}} & 36  & 619  & 158  & 9,926  & 48,901 & 0.1015  & 0.8985 \\
\multicolumn{1}{r|}{\begin{tabular}[c]{@{}r@{}}Health and Personal \\ Care\end{tabular}} & 39  & 627  & 180  & 11,489 & 56,430 & 0.1018  & 0.8982 \\
\multicolumn{1}{r|}{Home and Kitchen}  & 39  & 475  & 18 & 972 & 4,275 & 0.1137  & 0.8863 \\
\multicolumn{1}{r|}{Kindle Store} & 61  & 663  & 15 & 1,099  & 4,972 & 0.1105  & 0.8895 \\
\multicolumn{1}{r|}{Movies and TV}  & 109 & 1,319 & 149  & 20,071 & 98,265 & 0.1021  & 0.8979 \\
\multicolumn{1}{r|}{Musical Instruments} & 18  & 220  & 14 & 383 & 1,540 & 0.1244  & 0.8756 \\
\multicolumn{1}{r|}{Office Products} & 28  & 372  & 89 & 3,553  & 16554.0  & 0.1073  & 0.8927 \\
\multicolumn{1}{r|}{\begin{tabular}[c]{@{}r@{}}Patio Lawn and\\ Garden\end{tabular}}  & 18  & 174  & 31 & 797 & 2,697 & 0.1478  & 0.8522 \\
\multicolumn{1}{r|}{Pet Supplies} & 26  & 358  & 22 & 810 & 3,938 & 0.1028  & 0.8972 \\
\multicolumn{1}{r|}{Sports and Outdoor} & 33  & 424  & 17 & 781 & 3,604 & 0.1084  & 0.8916 \\
\multicolumn{1}{r|}{\begin{tabular}[c]{@{}r@{}}Tools and Home\\ Improvement\end{tabular}} & 27  & 276  & 10 & 344 & 1,380 & 0.1246  & 0.8754 \\
\multicolumn{1}{r|}{Toys and Games} & 31  & 347  & 25 & 906 & 4,337 & 0.1044  & 0.8956 \\
\multicolumn{1}{r|}{Video Games} & 37  & 408  & 29 & 1,229  & 5,916 & 0.1039  & 0.8961 \\ \bottomrule
\end{tabular}
\end{table*}

\bibliographystyle{IEEEtran}
\bibliography{references.bib}